\documentclass[11pt]{article}

\usepackage[letterpaper,margin=1in]{geometry}

\usepackage[T1]{fontenc}
\usepackage{mathptmx}
\usepackage{microtype}

\DeclareUnicodeCharacter{2191}{\ensuremath{\uparrow}}
\DeclareUnicodeCharacter{2193}{\ensuremath{\downarrow}}

\usepackage{amsmath}
\usepackage{amssymb}
\usepackage{mathtools}
\usepackage{bm}
\usepackage{bbm}        % \mathbbm{1}

\usepackage{graphicx}
\usepackage{booktabs}
\usepackage{multirow}
\usepackage[table]{xcolor}
\usepackage{siunitx}
\usepackage{csvsimple}

\usepackage{algorithm}
\usepackage[noend]{algpseudocode}

\usepackage{enumitem}
\usepackage{tikz}

\usepackage{eccvabbrv}

\definecolor{cvprblue}{rgb}{0.21,0.49,0.74}
\definecolor{lowexpression}{rgb}{0.8,0.8,0.8}

\usepackage[colorlinks,allcolors=cvprblue,breaklinks]{hyperref}
\usepackage{cleveref}
\usepackage{orcidlink}

\providecommand{\keywords}[1]{\par\medskip\noindent\textbf{Keywords:} #1}

\newcommand{\method}{HoloTea\xspace}

\title{3D-Guided Scalable Flow Matching for Generating Volumetric Tissue
Spatial Transcriptomics from Serial Histology}

\author{%
\small
\begin{minipage}{\dimexpr\textwidth-2\tabcolsep\relax}\centering
Mohammad Vali Sanian\textsuperscript{1,2,7*}, Arshia Hemmat\textsuperscript{6,7*}, Amirhossein Vahidi\textsuperscript{4,7*}, Jonas Maaskola\textsuperscript{7}, Jimmy Tsz Hang Lee\textsuperscript{7}, Martin Prete\textsuperscript{7}, Stanislaw Makarchuk\textsuperscript{7}, Yeliz Demirci\textsuperscript{7}, Nana-Jane Chipampe\textsuperscript{7}, Muzlifah Haniffa\textsuperscript{4,5,7}, Omer Bayraktar\textsuperscript{7\dag}, Lassi Paavolainen\textsuperscript{2\dag}, Mohammad Lotfollahi\textsuperscript{3,4,7\dag}
\\[4pt]
{\footnotesize
\textsuperscript{1}Department of Computer Science, University of Helsinki, Helsinki, Finland\quad
\textsuperscript{2}Institute for Molecular Medicine Finland (FIMM), University of Helsinki, Helsinki, Finland\quad
\textsuperscript{3}Cambridge Stem Cell Institute, University of Cambridge, Cambridge, UK\quad
\textsuperscript{4}Cambridge Centre for AI in Medicine, University of Cambridge, Cambridge, UK\quad
\textsuperscript{5}Department of Medicine, University of Cambridge, Cambridge, UK\quad
\textsuperscript{6}Department of Computer Science, University of Oxford, Oxford, UK\quad
\textsuperscript{7}Wellcome Sanger Institute, Hinxton, Cambridge, UK
\\[3pt]
\textsuperscript{*}Equal contribution.\quad \textsuperscript{\dag}Correspondence: \texttt{ob5@sanger.ac.uk}, \texttt{lassi.paavolainen@helsinki.fi}, \texttt{ml19@sanger.ac.uk}}
\end{minipage}%
}
\date{}

\begin{document}

\maketitle

\begin{abstract}
A 3D transcriptomic profile of a tissue can give a holistic view of its organization in health and disease, but spatial transcriptomics (ST) is costly, so usually only some serial sections are profiled, while H\&E is available for all. Most methods that infer ST from histology treat each section independently and ignore 3D structure, while existing 3D-aware approaches are not generative and do not scale well. We present Holographic Tissue Expression Inpainting and Analysis (\method), a 3D-aware flow-matching framework that imputes spot-level gene expression on an unprofiled section from its H\&E and the measured expression of adjacent profiled sections. Our key idea is to take the measured expression at the same registered position on neighboring sections and feed this cross-section context to the flow model through a lightweight ControlNet, so that predictions follow anatomical continuity along $z$. To capture the discrete, zero-inflated nature of counts, the flow starts from a learned zero-inflated negative binomial (ZINB) prior that, on an unprofiled section, is predicted from the measured expression of the nearest profiled section. A global attention block with a small set of inducing tokens adds whole-slide context at a cost linear in the number of spots, so training and inference scale to large 3D ST datasets. Across three ST datasets spanning different tissue types and resolutions, \method consistently improves 3D expression accuracy and generalization compared to 2D and 3D baselines. We envision \method advancing the creation of accurate 3D virtual tissues, ultimately accelerating biomarker discovery and deepening our understanding of disease. Code is available at \url{https://github.com/Lotfollahi-lab/HoloTea}.

\keywords{Spatial transcriptomics, Flow matching, 3D-guided Imputation}
\end{abstract}
\section{Introduction}

Spatial transcriptomics (ST)~\cite{Williams2022} links gene expression to tissue morphology and has transformed how we study cellular states and disease-associated niches~\cite{McGregor2025, Kanemaru2023} in situ~\cite{Zhang2022}. Existing platforms (e.g., Visium, Xenium, CosMx) operate on 2D sections with a limited field of view, capturing only a thin slice of a much larger 3D tissue volume, which restricts the study of vertical cell–cell communication, 3D niches, and morphomolecular architectures spanning multiple planes. Native 3D or thick-tissue in situ assays and 3D molecular atlases partially address this~\cite{Fang_2024,Sui2024-ct,doi:10.1126/science.adq2084} but remain constrained by capture depth, long protocols, specialized instrumentation, and cost; densely sectioning tissue and profiling many serial sections~\cite{Oliveira2025,Mo2024} is likewise prohibitively expensive and labor-intensive for routine 3D reconstruction.

In contrast, acquiring histology is cheap and scalable: tissues can be densely sectioned, H\&E-stained, and imaged as serial whole-slide stacks, with ST performed only on a subset of sections. This naturally motivates computational models that leverage these dense histology stacks and sparse ST measurements to learn the 3D spatial structure of gene expression. By predicting 3D-consistent spatial transcriptomics from serial H\&E, such models can reconstruct volumetric tissue molecular architecture without sequencing every section, enabling cost-effective, scalable analysis of 3D tissue niches and cell–cell communication in health and disease.

Within this emerging paradigm of H\&E-based ST prediction, two main streams of work have appeared. The first focuses on whole-slide generative modeling with flow matching, exemplified by STFlow~\cite{HuangSTFlow}, which models the joint distribution of all spots on a slide using local spatial attention and efficient slide-level encoders. The second is 3D-aware imputation, exemplified by ASIGN~\cite{Zhu_2025_CVPR_ASIGN}, which propagates expression from one ST-profiled section across a serial stack with an anatomy-aware graph network; it is not generative and its memory grows with the number of slides and spots. While effective, these approaches leave open two key gaps for 3D reconstruction from serial histology: (i) most H\&E to ST predictors are trained per section and do not explicitly enforce 3D coherence across serial stacks, leading to drift along the $z$-axis; and (ii) the start distribution used by flow-matching models is rarely tailored to the zero-inflated, overdispersed nature of gene counts ~\cite{Lopez2022, lopez2018scvi}, placing undue burden on the denoiser early in the generative trajectory.

% We address both issues with \method, a 3D H\&E to ST generation built on flow matching~\cite{lipman2022flowmatching,liu2023rectifiedflow}. For each query spot on section $z$, we retrieve morphologically corresponding candidates on $z{\pm}1$ using $(x,y)$ proximity and feature-space cosine similarity, then inject this adjacent context into selected denoiser blocks via a lightweight ControlNet~\cite{zhang2023controlnet}; this registration-free design promotes anatomical continuity across sections without heavy preprocessing. Because FM decouples the probability path from the start distribution, we instantiate biology-compatible priors: a learned ZINB prior with the scVI parameterization~\cite{lopez2018scvi,gayoso2022scvitools} and a $z$-aware spatial-empirical prior. Compared to STFlow~\cite{HuangSTFlow}, our contributions are orthogonal and complementary: we retain FM stability and slide-level efficiency while adding explicit 3D conditioning and priors aligned to transcript count statistics. Compared to ST-Assign~\cite{stassign2023}, our H\&E-only generator avoids reference coupling and global mixture inference, yielding a training/inference cost that is linear in the number of spots (local kNN graphs plus $O(Nm)$ inducing attention) and naturally scales to whole slides and serial stacks. An overview of the retrieval-conditioned flow-matching pipeline, including \(z{\pm}1\) neighbor search, adjacent-token aggregation, lightweight ControlNet conditioning, and ZINB priors, is shown in Fig.~\ref{fig:adjacent}.

\begin{figure*}[t]
    \centering
    \includegraphics[width=\linewidth]{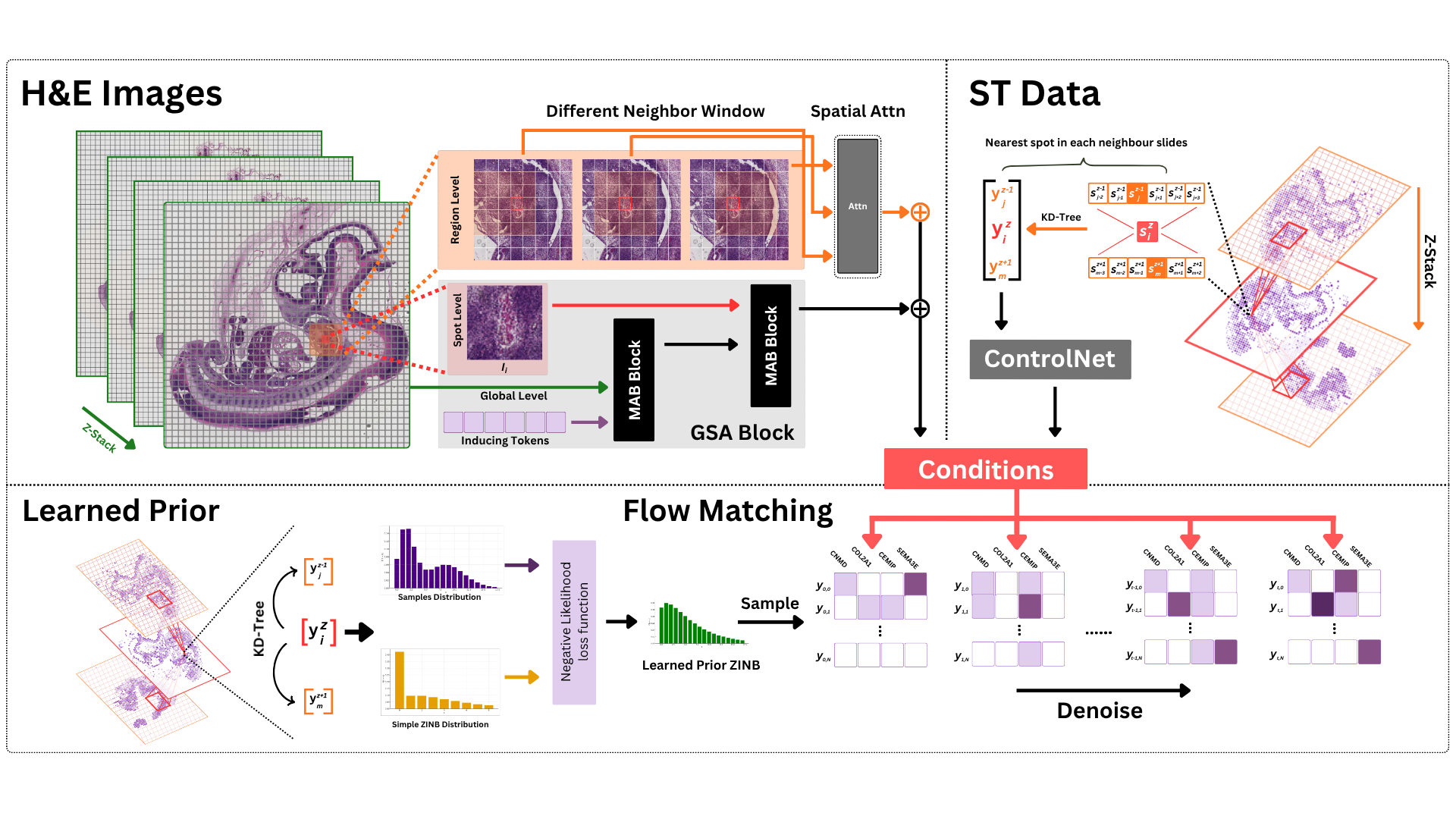}
    \caption{Method overview. For each query spot on section $z$, \method takes the nearest spot by position on sections $z{\pm}1$ as adjacent tokens and uses a ControlNet to condition the flow-matching denoiser with ZINB-based priors.}
    \label{fig:adjacent}
\end{figure*}

To address these gaps, we introduce \method, a 3D H\&E$\to$ST generator built on flow matching~\cite{lipman2022flowmatching,liu2023rectifiedflow}. For each query spot on section $z$, we take the spatially corresponding spot, the nearest by $(a,b)$, on the registered sections $z{\pm}1$, and inject its measured expression into the denoiser via a lightweight ControlNet~\cite{zhang2023controlnet}, promoting anatomical continuity. Because flow matching decouples the probability path from the start distribution, we instantiate a biology-compatible prior: a learned ZINB prior using the scVI parameterization~\cite{lopez2018scvi,gayoso2022scvitools}, predicted on an unprofiled section from the measured expression of the nearest profiled section. Unlike STFlow~\cite{HuangSTFlow}, which operates on single 2D sections, we extend flow-matching ST generation to the serial-section regime; unlike ASIGN~\cite{Zhu_2025_CVPR_ASIGN}, a deterministic graph regressor whose memory grows with the stack, our model is generative, with training and inference cost linear in the number of spots (local kNN graphs plus $O(Nm)$ inducing-point attention), scaling to whole slides and serial stacks. \cref{fig:adjacent} overviews the pipeline.
% Compared to STFlow~\cite{HuangSTFlow}, our contributions are orthogonal and complementary: we retain FM stability and slide-level efficiency while adding explicit 3D conditioning and priors aligned to transcript count statistics. Compared to ST-Assign~\cite{stassign2023}, our H\&E-only generator avoids reference coupling and global mixture inference, yielding a training/inference cost that is linear in the number of spots (local kNN graphs plus $O(Nm)$ inducing attention) and naturally scales to whole slides and serial stacks. An overview of the retrieval-conditioned flow-matching pipeline, including \(z{\pm}1\) neighbor search, adjacent-token aggregation, lightweight ControlNet conditioning, and ZINB priors, is shown in Fig.~\ref{fig:adjacent}.

\paragraph{Our contributions are:} 
\begin{itemize}
    \item An adjacent-slide conditioning scheme that retrieves spatially matched context from adjacent registered sections and injects it with a spot-wise ControlNet.  
    \item A cross-section-informed, biology-compatible start distribution for flow matching: a learned ZINB prior.
    \item A stable, scalable global attention to inject the global image features that scales linearly with the number of spots without scaling with the number of slides.  
    \item We demonstrate the strong performance of our method across multiple datasets for whole slide gene expression generation task: \method shows competitive results compared to other baselines.
\end{itemize}

\section{Related Work}

\subsection{H\&E to Expression prediction}
Large-scale H\&E$\to$ST models have advanced along three axes: stronger multi-resolution encoders, better supervision/benchmarks, and generative objectives. Discriminative regressors include TRIPLEX, which fuses cellular-, meso-, and slide-scale cues for consistent gains on Visium~\cite{Chung2024TRIPLEX}; DANet, which uses dynamic cross-modal alignment while accounting for gene–gene dependencies~\cite{Wu2025DANet}; and mclSTExp, which combines transformer attention with image–omics contrastive learning~\cite{Min2024mclSTExp}. SpaRED curates a 26-source benchmark with harmonized preprocessing and shows that completing corrupted ST matrices significantly boosts downstream predictors~\cite{Mejia2024SpaRED}. Beyond discriminative regression, generative approaches treat ST as conditional generation: Stem uses diffusion to capture spot-level stochasticity given histology and reports SOTA across several datasets~\cite{Zhu2025DiffSGE}. Pathology foundation encoders (UNI/UNI2), pretrained on $100$M–$200$M patches, are increasingly used as robust backbones for H\&E feature extraction in ST pipelines~\cite{UNI_NatMed2024}.

\subsection{3D spatial transcriptomics and 3D imputation}
For imputation from serial histology, ASIGN~\cite{Zhu_2025_CVPR_ASIGN} registers a stack of H\&E sections, builds an anatomy-aware graph across them and predicts expression on all sections from a single ST-profiled one; we use it as our 3D baseline. On the data and systems side, Open-ST provides end-to-end protocols and tooling for subcellular ST in 2D/3D, including serial-section registration and 3D ``virtual tissue block'' reconstruction~\cite{OpenST_Cell_2024,OpenST_StarProt_2025}, and SpaDo contributes cross-slice spatial-domain detection, reference-based annotation, and clustering across platforms~\cite{Duan2024SpaDo}. Larger multimodal corpora also pair WSIs with ST at scale: HEST-1k consolidates $1{,}229$ paired ST and H\&E samples (2.1M expression–morphology pairs) with a standardized ingestion and benchmarking library, facilitating stack-aware evaluation and the use of strong pathology backbones for volumetric imputation~\cite{HEST2024}.

% \title{STFlow-3D: 3D-Aware Flow-Matching Imputation of Spatial Transcriptomics from H\&E}
\section{Preliminary}

\subsection{Problem setup}
We consider serial sections indexed by $z\in\{1,\ldots,Z\}$ with spot sets $\mathcal{S}_z$; each spot $i\in\mathcal{S}_z$ has coordinates and section index $\mathbf{s}_i=(a_i,b_i,z_i)$, an H\&E tile $I_i$, a target gene-expression vector $\mathbf{y}_i\in\mathbb{R}_{\ge 0}^{G}$ (raw counts or $\log(1{+}\cdot)$), and an image embedding $\mathbf{v}_i=\phi(I_i)\in\mathbb{R}^{D}$ from a pretrained encoder $\phi(\cdot)$; we use the frozen UNI pathology foundation model~\cite{UNI_NatMed2024} ($224{\times}224$ tiles, $D{=}1024$; other encoders in \cref{tab:s-encoders}). We aim to impute $\hat{\mathbf{y}}_i$ from $(I_i,\mathbf{s}_i)$ with high within-section fidelity and across-$z$ coherence, building a within-section $k$-nearest-neighbor (k-NN) graph in $(a,b)$.

\subsection{Background}
\paragraph{Flow matching.}
We model imputation as transport from a start distribution $p_0$ over gene vectors to the data distribution $p_1$ via a time-dependent field learned with flow matching (FM). For each spot $i$, we draw a start sample $\mathbf{x}_{0,i}\!\sim\!p_0$ and define a linear (displacement) path to the target $\mathbf{x}_{1,i}\equiv\mathbf{y}_i$,
\begin{equation}\label{eq:fm-interp}
\mathbf{x}_{t,i}=(1-t)\,\mathbf{x}_{0,i}+t\,\mathbf{x}_{1,i},\quad t\in[0,1],
\end{equation}
where $t$ is the scalar time along the path. A time-conditioned network $f_\theta$ (parameters $\theta$) predicts the clean endpoint from partial information,
\begin{equation}
\hat{\mathbf{y}}_i=f_\theta\!\big(\mathbf{x}_{t,i},\,\mathbf{c}_i,\,t\big),\quad 
\mathbf{c}_i=[\mathbf{v}_i,\,\mathbf{s}_i,\,\mathrm{kNN}(i)],
\end{equation}
where $\mathbf{c}_i$ concatenates the image features $\mathbf{v}_i$, the coordinates $\mathbf{s}_i$, and a neighborhood summary $\mathrm{kNN}(i)$ (the indices or pooled features of the $k$ nearest spots to $i$ within its section). We minimize the supervised endpoint-regression loss
$\mathcal{L}_\mathrm{fm}=\mathbb{E}_{i,t}\big[\|\hat{\mathbf{y}}_i-\mathbf{y}_i\|_2^2\big]$\label{eq:fm-loss}, following standard FM practice for displacement or straight paths~\cite{lipman2022flowmatching,liu2023rectifiedflow,esser2024rectified}.

\section{Method}
Our method contains three main components: (i) conditioning on the measured expression of the matched spots on the closest profiled sections, together with a learned ZINB prior trained on the training sections; (ii) applying conditioning and $z$-variation through ControlNet~\cite{zhang2023controlnet}; and (iii) building hierarchical feature selection efficiently via Global set attention (GSA)~\cite{lee2019settransformer}.

\subsection{Adjacent-section conditioning}\label{sec:adj}
For a query spot $i\in\mathcal{S}_z$, we take the spatially corresponding spot on each adjacent profiled section of the registered stack,
\begin{equation}\label{eq:adj-candidates}
j^{\pm}(i)=\arg\min_{j\in\mathcal{S}_{z\pm1}}\big\|(a_i,b_i)-(a_j,b_j)\big\|,
\end{equation}
where $(a_i,b_i)$ are the planar coordinates of spot $i$ and $\mathcal{S}_{z\pm1}$ are the spot sets on the adjacent sections $z\!\pm\!1$ in the training data. Here $z\!\pm\!1$ denotes the nearest profiled sections below and above $z$. During training, held-out sections are absent, so under the even-slice hold-out these are the next profiled sections ($\Delta z{=}2$); at inference, a held-out section's nearest profiled sections are its immediate neighbours ($\Delta z{=}1$). Because all sections are registered into a common frame (\cref{app:spatial-alignment}), $j^{\pm}(i)$ lies at the same anatomical position as $i$. The adjacent tokens are the measured expression of these spots, log-normalized as the targets,
\begin{equation}\label{eq:adj-token-train}
\tilde{\mathbf{g}}^{\pm}_i=\mathbf{y}_{j^{\pm}(i)}\in\mathbb{R}^{G},
\end{equation}
one for the section below and one for the section above. At inference, \cref{eq:adj-token-train} is used unchanged: $\mathbf{y}_{j^{\pm}(i)}$ is the measured expression of the held-out section's nearest profiled sections, which also feed the learned prior (\cref{sec:prior}). Neighbour states are not generated jointly, and the held-out section's own expression is never used.

\paragraph{ControlNet injection.}
ControlNet turns per-spot gene vectors into per-spot control tokens: it projects each spot's vector to a few channels and places it at the spot's position $(a_i,b_i)$ on a coarse 2D grid (a \emph{gene map}), applies a small CNN so that each location sees its spatial neighbourhood, and reads the result back at $(a_i,b_i)$ by bilinear interpolation. Applied separately to each adjacent token $\tilde{\mathbf{g}}^{\pm}_i$ (and, in a separate branch, to the current flow state $\mathbf{x}_t$), this gives
\begin{equation}\label{eq:controlnet-u}
\mathbf{u}^{(b)}_i=h^{(b)}_\omega\big(\tilde{\mathbf{g}}^{(b)}_i\big)\in\mathbb{R}^{d_u},\qquad b\in\{+,-,\mathbf{x}_t\},
\end{equation}
where $h^{(b)}_\omega$ is the projection, CNN and read-out above with separate weights per branch ($\tilde{\mathbf{g}}^{(\mathbf{x}_t)}_i=\mathbf{x}_{t,i}$), and $d_u$ is the control-token dimensionality. These tokens are added once to the input tokens of the backbone,
\begin{equation}\label{eq:controlnet-inject}
\mathbf{h}^{(0)}_i=\mathbf{W}\mathbf{v}_i+\gamma(t)+\alpha(t)\textstyle\sum_b \mathbf{u}^{(b)}_i,\qquad \alpha(t)=t\,w,
\end{equation}
where $\gamma(t)$ is a Fourier time embedding, $b$ runs over the ControlNet branches (adjacent section above, below, and $\mathbf{x}_t$), and $w$ ramps from 0 to 1 over the first training epochs; scaling by $t$ lets control grow as the flow approaches the data, avoiding early over-steering, adapting ControlNet~\cite{zhang2023controlnet} to spot-wise conditioning.

\subsection{Cross-section learned ZINB prior}\label{sec:prior}
\paragraph{Learned ZINB prior.}
We \emph{pretrain} a predictor $\psi_\eta$ that maps a spot's measured log-counts to per-spot ZINB parameters,
\begin{equation}\label{eq:zinb-map}
\psi_\eta:\ \log\!\big(1+\mathbf{c}_i\big)\ \mapsto\ (\boldsymbol{\mu}_i,\boldsymbol{\pi}_i)\in\mathbb{R}_{>0}^{G}\times(0,1)^{G},
\end{equation}
together with a per-gene inverse dispersion $\boldsymbol{\theta}\in\mathbb{R}_{>0}^{G}$ shared across spots. Here $\mathbf{c}_i\in\mathbb{R}_{\ge 0}^{G}$ are measured counts: during pretraining and flow-matching training, $\mathbf{c}_i=\mathbf{y}_i$ (the spot's own counts); at inference on an unprofiled section, $\mathbf{c}_i=\mathbf{y}_{j^\ast(i)}$, the counts of the spot $j^\ast(i)$ nearest to $(a_i,b_i)$ on the closest profiled training section along $z$. Held-out expression is therefore never used, and measurements from adjacent profiled sections enter the model through the flow start distribution. Concretely, $\psi_\eta$ encodes $\log(1+\mathbf{c}_i)$ into a spot embedding $\mathbf{e}_i$, adds a learned gene embedding $\mathbf{g}_g$, and decodes $\mu_{i,g}$ and $\pi_{i,g}$ from $\mathbf{e}_i+\mathbf{g}_g$ (softplus and sigmoid outputs) and $\theta_g$ from $\mathbf{g}_g$ alone (softplus). $\boldsymbol{\mu}_i$ are per-gene means, $\boldsymbol{\theta}$ inverse-dispersions, and $\boldsymbol{\pi}_i$ zero-inflation probabilities (scVI parameterization~\cite{lopez2018scvi,gayoso2022scvitools}; see ZINB in \cref{pre:zinb}). The pretraining minimizes the negative log-likelihood of the ZINB model on the training expressions:
\begin{equation}\label{eq:zinb-nll}
\mathcal{L}_{\mathrm{ZINB}}(\eta)= - \sum_{i\in\mathcal{D}_{\mathrm{train}}}\sum_{g=1}^{G}\log p_{\mathrm{ZINB}}\!\big(y_{i,g}\,\big|\,\mu_{i,g},\theta_{g},\pi_{i,g}\big),
\end{equation}
with $(\boldsymbol{\mu}_i,\boldsymbol{\pi}_i)=\psi_\eta\big(\log(1+\mathbf{y}_i)\big)$. After convergence we \emph{freeze} $\psi_\eta$ and use it solely to define the flow start: sample counts
\begin{equation}\label{eq:zinb-sample}
\mathbf{Y}_i \sim \prod_{g=1}^G \mathrm{ZINB}(\mu_{i,g},\theta_{g},\pi_{i,g}),\qquad
\mathbf{x}_{0,i}=\log\!\big(1+\mathbf{Y}_i\big).
\end{equation}
No gradients from the flow-matching loss backpropagate into $\psi_\eta$; the start distribution $p_0$ is therefore fixed during FM training. STFlow~\cite{HuangSTFlow} also starts from a ZINB, but with fixed parameters shared by all spots and genes; ours is per spot and per gene and is predicted from measured counts of the nearest profiled section, so the flow starts close to the matched neighbouring spot's expression. \cref{tab:ablation_component} adds this prior, ControlNet and the cross-slide branch to STFlow one at a time, and the $m$-sweep (\cref{tab:m_sweep_embryo_pow2}; $m{=}0$ is STFlow) isolates global attention.

\subsection{Global set attention (GSA)}
To inject global context without quadratic cost in the number of spots $N$, we use a two-stage attention block inspired by the Induced Set Attention Block (ISAB)~\cite{lee2019settransformer}. Let $\mathbf{X}\!\in\!\mathbb{R}^{N\times d}$ be spot tokens (per spot: current state $\mathbf{x}_{t,i}$, H\&E features, time/positional embeddings) and let $\mathbf{S}\!\in\!\mathbb{R}^{m\times d}$ be a small set of inducing tokens ($m\!\ll\! N$), learned and shared across batches. We use multi-head attention $\mathrm{MHA}(\mathbf{Q},\mathbf{K},\mathbf{V})$ with $h$ heads.

Stage 1 (global read): the inducing tokens query all spots to form a global summary
\begin{equation}\label{eq:gsa-read}
\mathbf{H} \;=\; \mathrm{MHA}\!\big(\mathbf{S},\,\mathbf{X},\,\mathbf{X}\big) \;\in\; \mathbb{R}^{m\times d},
\end{equation}
which aggregates whole-slide information into $m$ slots. Stage 2 (global write): each spot then attends to this compact summary
\begin{equation}\label{eq:gsa-write}
\mathbf{Y} \;=\; \mathrm{MHA}\!\big(\mathbf{X},\,\mathbf{H},\,\mathbf{H}\big) \;\in\; \mathbb{R}^{N\times d},
\end{equation}
injecting global context back into all spots. We wrap both stages in pre-norm residual blocks followed by a position-wise MLP with residual. The cost scales as $O(Nm\,d)$ rather than $O(N^2 d)$, propagating whole-slide information without attention cost growing with the number of spots; we use $m{=}64$ for all datasets, making GSA a drop-in global-context module for whole-slide and 3D stacks.

\subsection{Inference}
We impute expression without ground-truth $\mathbf{y}$ by marching the flow from $t{=}0$ to $t{=}1$ on a grid $\{t_s\}_{s=0}^{S}$ ($t_0{=}0$, $t_S{=}1$, step weights $\eta_s{=}(t_{s+1}{-}t_s)/(1{-}t_s)$), reusing adjacent-section context. We (1)~compute $\mathbf{v}_i{=}\phi(I_i)$ and sample a start $\mathbf{x}_{0,i}$ from the learned ZINB prior $\psi_\eta$; (2)~form the adjacent tokens $\tilde{\mathbf{g}}^{\pm}_i$ once from the measured expression of the nearest profiled sections (\cref{eq:adj-token-train}), and at each step derive ControlNet tokens $\mathbf{u}_i$ and run the Spatial Transformer with GSA to obtain $\hat{\mathbf{y}}^{(s)}_i$; and (3)~advance along the straight path toward the prediction,
\begin{equation}
\mathbf{x}_{t_{s+1},i} \;\leftarrow\; (1-\eta_s)\,\mathbf{x}_{t_s,i} \;+\; \eta_s\,\hat{\mathbf{y}}^{(s)}_i.
\end{equation}
We report the last-step prediction $\hat{\mathbf{y}}_i$ of a single trajectory; since $f_\theta$ is trained with squared error, this is a point estimate (conditional mean), not one plausible sample (\cref{supp:point-estimate}).

\section{Experiments}
\subsection{Experiments settings}

\sisetup{
  separate-uncertainty = true,
  uncertainty-separator = \pm,
  table-align-uncertainty = true,
  detect-weight = true,
  detect-inline-weight = math
}

\begin{table}[t]
  \centering
  \scriptsize
  \setlength{\tabcolsep}{5pt}
  \renewcommand{\arraystretch}{1.15}
  \caption{\textbf{Gene-expression imputation on held-out sections (HVG-250 panel).}
    HER2 and ST-Data use split (i) (one ST-labeled section per stack), averaged over its four folds; the in-house embryo uses split (ii) (even-slice hold-out).
    PCC SPOT is the Pearson correlation across genes within each spot, averaged over spots; PCC GENE is the Pearson correlation across spots for each gene, averaged over genes.
    $\pm$ is the standard deviation across spots (PCC SPOT) or across genes (PCC GENE) within an evaluation, averaged over folds; it is not a variation across splits.
    \textbf{Our Method} (\method) is the full model: learned ZINB prior (frozen, defines the flow start), ControlNet conditioning and the cross-slide branch. STFlow uses a fixed global ZINB prior without these components and shares \method's backbone and hyperparameters; UNI, CIGA and GigaPath are image-encoder baselines trained with the same protocol (\cref{app:experiments_details}). Best results in bold.}

  \resizebox{\textwidth}{!}{%
  \begin{tabular}{
    l
    S[table-format=1.3]
    S[table-format=1.3]
    S[table-format=1.3(3)]
    S[table-format=1.3(3)]
    S[table-format=1.3]
    S[table-format=1.3]
    S[table-format=1.3(3)]
    S[table-format=1.3(3)]
    S[table-format=1.3]
    S[table-format=1.3]
    S[table-format=1.3(3)]
    S[table-format=1.3(3)]
  }
    \toprule
    & \multicolumn{4}{c}{\textit{HER2}}
    & \multicolumn{4}{c}{\textit{ST-Data (images)}}
    & \multicolumn{4}{c}{\textit{In-house embryo}} \\
    \cmidrule(lr){2-5}\cmidrule(lr){6-9}\cmidrule(lr){10-13}
    \textbf{Model}
      & \multicolumn{1}{c}{\textbf{MSE} (↓)} & \multicolumn{1}{c}{\textbf{MAE} (↓)} & \multicolumn{2}{c}{\textbf{PCC} (↑)}
      & \multicolumn{1}{c}{\textbf{MSE} (↓)} & \multicolumn{1}{c}{\textbf{MAE} (↓)} & \multicolumn{2}{c}{\textbf{PCC} (↑)}
      & \multicolumn{1}{c}{\textbf{MSE} (↓)} & \multicolumn{1}{c}{\textbf{MAE} (↓)} & \multicolumn{2}{c}{\textbf{PCC} (↑)} \\
    \cmidrule(lr){4-5}\cmidrule(lr){8-9}\cmidrule(lr){12-13}
      & & & \multicolumn{1}{c}{\textbf{SPOT}} & \multicolumn{1}{c}{\textbf{GENE}}
      & & & \multicolumn{1}{c}{\textbf{SPOT}} & \multicolumn{1}{c}{\textbf{GENE}}
      & & & \multicolumn{1}{c}{\textbf{SPOT}} & \multicolumn{1}{c}{\textbf{GENE}} \\
    \midrule

    UNI
      & 1.093 & 0.859 & 0.509(94)  & 0.259(231)
      & 0.769 & 0.699 & 0.569(77)  & 0.332(149)
      & 0.261 & 0.296 & 0.506(224) & 0.613(203) \\

    CIGA
      & 0.865  & 0.738 & 0.494(135) & 0.465(143)
      & 0.783 & 0.706 & 0.541(83)  & 0.247(137)
      & 0.449  & 0.438 & 0.399(194) & 0.427(229) \\

    Gigapath
      & 0.611 & 0.615 & 0.581(126) & 0.576(161)
      & 0.674 & 0.649 & 0.597(76)  & 0.364(145)
      & 0.314 & 0.322 & 0.457(223) & 0.554(225) \\

    STFlow
      & 0.477 & 0.534 & 0.740(125) & 0.620(146)
      & 0.443 & 0.532 & 0.775(80)  & 0.540(142)
      & 0.255 & 0.270 & 0.557(248) & 0.663(166) \\

    \cmidrule(lr){1-13}
    \textsc{ASIGN} 3D
      & 0.462 & 0.563 & 0.704(113) & 0.607(112)
      & 0.481 & 0.551 & 0.741(13)  & 0.531(121)
      & 0.623 & 0.634 & 0.343(191) & 0.453(120) \\

    \rowcolor{cyan!10}
    \textbf{Our Method}
      & \bfseries 0.455 & \bfseries 0.523 & \bfseries 0.744(114) & \bfseries 0.638(145)
      & \bfseries 0.364 & \bfseries 0.468 & \bfseries 0.796(73)  & \bfseries 0.612(132)
      & \bfseries 0.199 & \bfseries 0.235 & \bfseries 0.576(245) & \bfseries 0.691(173) \\

    \bottomrule
  \end{tabular}%
  }
  \label{tab:three_datasets_metrics}
\end{table}
\paragraph{Datasets.}
We evaluate on two public benchmarks  the \textbf{HER2} multi-section breast cancer dataset~\cite{Andersson2021HER2NatComm} and the \textbf{ST-Data} cohort introduced with ST-Net~\cite{He2020STNet}  following the authors’ preprocessing with section-level holdouts (entire held-out sections). We additionally benchmark on an \textbf{in-house Embryo dataset} (to be released in a separate manuscript): a 3D stack of 52 serial embryo sections profiled with 10x Visium, providing a dense multi-modal volume (H\&E\,$+$\,ST) co-registered across the stack.

For ASIGN~3D \cite{Zhu_2025_CVPR_ASIGN}, we re-ran the method on the HER2 and ST datasets, as well as on our in-house embryo dataset, using the normalization described in \cite{HuangSTFlow}. This ensures a fair comparison of MAE and MSE. Hyperparameters are listed in Appendix \ref{app:experiments_details}.

\paragraph{Gene sets.}
\label{sec:imputing-marker-genes}
We report results on three panels: (i) \textbf{Top-250 HVG} selected with Scanpy’s highly-variable-genes routine (Seurat flavor)~\cite{wolf2018scanpy,Stuart2019SeuratV3};
(ii) a \textbf{Custom gene set} of $\sim$150 genes that are highly variable across the whole stack: per-section HVGs are pooled within 11 bins of sections spread along $z$, and the top 50 genes per bin are unioned;
 (iii) \textbf{marker genes} for 106 anatomical regions, identified by differential gene expression analysis; taking the top two markers per region yielded 211 distinct genes.

\paragraph{Evaluation metrics.}
We compute spot-wise Pearson correlation coefficient (PCC spot), gene-wise Pearson correlation coefficient (PCC gene) and MSE on held-out sections.

\subsection{Gene Expression Imputation}
We perform 3D spatial imputation as generating a whole slide gene expression from serial H\&E sections, given ground-truth ST on only a subset of sections. We adopt two complementary splits. 
{(i) Following the ASIGN 3D-ST task~\cite{Zhu_2025_CVPR_ASIGN}, each stack provides one ST-labeled section for supervision; all remaining sections are imputation targets, evaluated without using their ST during training.
(ii) Even-slice hold-out: within each stack we hold out even-index sections ($z{=}2,4,\dots$) as targets and train on the odd-index sections ($z{=}1,3,\dots$), probing across-$z$ generalization under fixed anatomy. For every held-out section, ground-truth ST is used only for evaluation: never for training, adjacent-section conditioning, prior construction, or inference. The only exception is the MOSCOT pre-registration of the in-house embryo, which uses all sections' ST to estimate the geometric alignment (rigid/affine transforms of coordinates and images); no expression values are transferred between sections (\cref{app:spatial-alignment}).

\cref{tab:three_datasets_metrics} reports one split per dataset, all on the HVG-250 panel: split (i) on the public HER2~\cite{Andersson2021HER2NatComm} and ST-Data~\cite{He2020STNet} breast datasets, averaged over its four folds, and split (ii) on the in-house embryo stack (simulating missing-through-plane acquisition at scale). \cref{tab:embryo_custom_panels} reports the embryo under the custom gene panel with the same split. \method outperforms all baselines across the three datasets (\cref{tab:three_datasets_metrics,tab:embryo_custom_panels}); ASIGN scales poorly on Embryo because its memory grows with the number of slides and spots (\cref{exp:comp}). We also compare with TRIPLEX~\cite{Chung2024TRIPLEX} on the public datasets (\cref{tab:triplex}).

\subsection{\method imputed expression preserves biological structure}
\label{sec:biological-validity}
\begin{figure*}[t]
  \centering
  \includegraphics[width=0.9\linewidth]{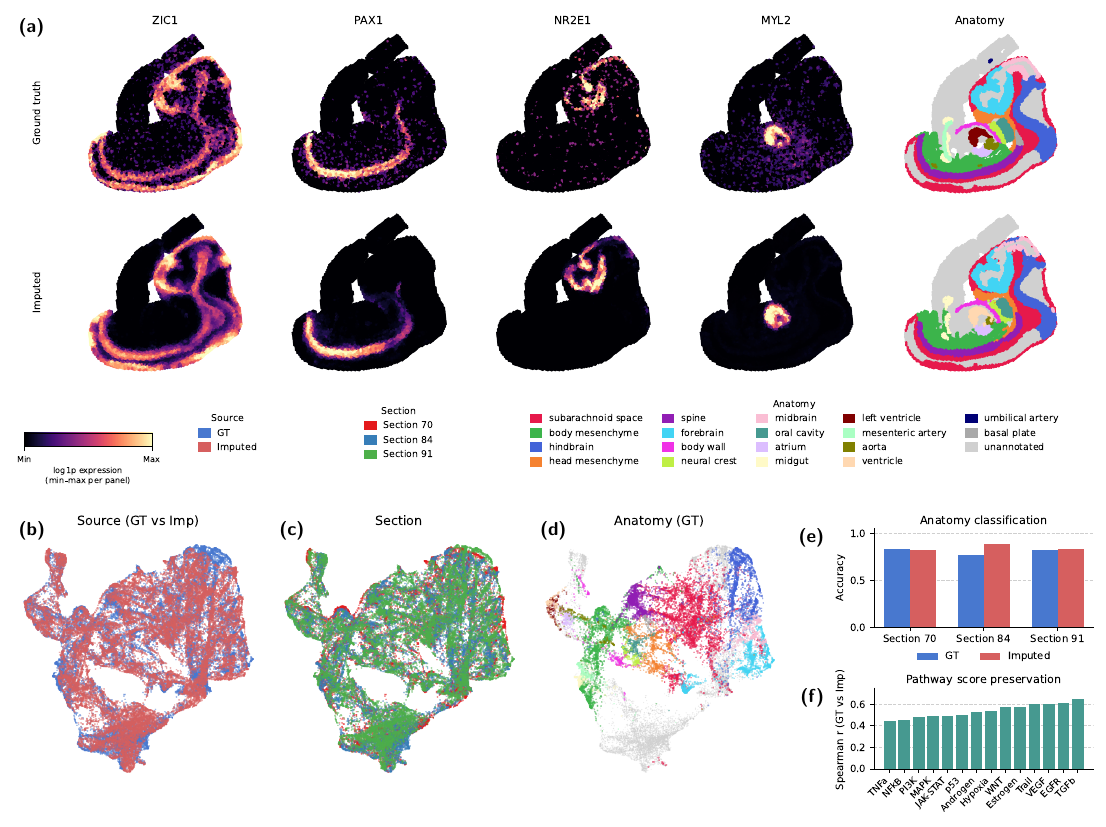}
  \caption{%
    \textbf{\method imputed expression preserves anatomical and biological structure.}
    (\textit{a})~Spatial expression of four marker genes on section~70 (GT top, imputed bottom); rightmost column: GT anatomical annotation (top) and classifier prediction from imputed expression (bottom).
    (\textit{b})--(\textit{d})~Joint UMAP of GT and imputed expression colored by source, section, and anatomical region.
    (\textit{e})~Leave-one-section-out anatomy classification accuracy (logistic regression).
    (\textit{f})~Spearman $r$ between GT and imputed PROGENy pathway scores (14 pathways, $n{=}167{,}584$ spots).
    See \cref{suppl-fig:expression-grid,suppl-fig:region-accuracy,suppl-tab:classifier-metrics,suppl-fig:enrichment,suppl-fig:pathway-scatter} for extended results.
    \label{fig:biological-validity}
  }
\end{figure*}

Training on the 211 anatomical marker genes (\cref{sec:imputing-marker-genes}), we ask whether imputed expression recovers biologically meaningful organization beyond reconstruction metrics. Annotations are available for 3 of the 26 held-out sections (70, 84, 91); pathway evaluation needs none and covers all 26.

\paragraph{Spatial expression patterns.}
\cref{fig:biological-validity}(a) shows GT and imputed expression for four marker genes, ZIC1 (subarachnoid space), PAX1 (spine), NR2E1 (forebrain), and MYL2 (left ventricle), selected as top regional markers by log2 fold-change (\cref{suppl-fig:expression-grid}).
Imputed expression closely recapitulates the known spatial distributions of all four genes while suppressing low-count shot noise.

\paragraph{Embedding structure.}
Joint PCA+UMAP of GT and imputed expression (\cref{fig:biological-validity}(b)--(d)) shows no systematic separation by source, confirming imputed expression occupies the same latent manifold as ground truth.

\paragraph{Anatomy classification and region-specific enrichment.}
Under leave-one-section-out CV, imputed expression preserves anatomical structure recoverable by linear classifiers: logistic regression matches or exceeds GT (overall accuracy $0.847$ vs.\ $0.807$; macro-F1 $0.657$ vs.\ $0.664$), whereas the non-linear $k$-NN and random-forest classifiers degrade on imputed expression (accuracy $0.716$ vs.\ $0.812$ and $0.782$ vs.\ $0.812$), most on section 91 (\cref{fig:biological-validity}(e); per-classifier and per-section results in \cref{suppl-tab:classifier-metrics}; \cref{suppl-fig:region-accuracy,suppl-fig:expression-grid}).
Per-gene log2 fold-change enrichment between each region and the rest yields a median Spearman $r{=}0.875$ (range $0.72$--$0.96$, 17 regions, \cref{suppl-fig:enrichment}), confirming that region-specific transcriptional signatures are faithfully preserved.

\paragraph{Pathway score preservation.}
PROGENy pathway scoring~\cite{Schubert2018,Badia2022} across all 26 held-out sections ($n{=}167{,}584$ spots) yields median Spearman $r{=}0.533$ between GT and imputed activity scores (14 pathways, range $0.45$--$0.65$, all $p{\approx}0$, \cref{fig:biological-validity}(f), \cref{suppl-fig:pathway-scatter}), demonstrating that \method preserves broad transcriptional signaling structure without any pathway supervision during training.

\subsection{Computational Complexity} \label{exp:comp}
Our method scales with the axial depth $z$: as the number of slides (and spots $N$) grows, \textsc{ASIGN}'s GPU usage rises sharply on our in-house dataset, whereas our footprint grows approximately linearly in $z$ during training, and inference needs only adjacent-section ($z\pm1$) features per target section (\cref{tab:complexity_spots_flops_mem}; per-sample FLOPs and peak training memory).

\section{Ablation}
We ablate four cumulative variants, (i)~Vanilla STFlow, (ii)~$+$\,learned ZINB prior, (iii)~$+$\,ControlNet conditioning, and (iv)~$+$\,cross-slide branch, on the in-house embryo even-slice hold-out across 50/100/200/500-gene panels (MSE, MAE, PCC; full table and per-panel analysis in \cref{tab:ablation_component} and \cref{app:extended-results}). Each component contributes: the learned ZINB prior helps most on medium panels; ControlNet conditioning improves PCC at every panel size and MSE at 50--200 genes over Vanilla STFlow and the prior-only model; and the cross-slide branch yields the strongest model overall (e.g., at 200 genes MSE $0.155\!\to\!0.147$ and PCC $0.694\!\to\!0.698$), indicating cross-slide cues add complementary long-range context.

\section{Conclusion}
We presented \method, a 3D-aware H\&E to ST generative model coupling flow matching with (i)~\textbf{3D context} via spatially corresponding spots from adjacent registered sections and ControlNet injection, (ii)~\textbf{generalization} through biology-compatible count priors, and (iii)~\textbf{scalability} via a transformer backbone with low-rank global context.
Across public breast datasets and a large in-house embryo stack, \method improves spot- and gene-wise correlation and reduces reconstruction error over prior H\&E to ST methods; ablations confirm complementary gains from the prior, ControlNet, and cross-slide branch, and imputed expression preserves biologically meaningful organization (anatomy, region signatures, pathway activity) while scaling near-linearly with stack depth.

\paragraph{Use case and future work.}
A typical setting is a densely sectioned H\&E stack in which only every few sections are profiled with ST. Imputing the unprofiled sections makes volume-level quantities measurable at every H\&E section rather than only at profiled ones, e.g.\ the 3D extent and boundaries of marker-defined structures such as the MYL2-positive left ventricle or the NR2E1-positive forebrain, and expression gradients between profiled sections, while fewer sections need costly profiling.
\method currently targets array-based ST on a spot grid. Extending it to single-cell, non-grid platforms such as Xenium~\cite{Luecken2025XeniumBestPractices} and CosMx, or to proteomic layers, requires cell positions on unprofiled sections; these can come from nuclei segmented on their H\&E, since the denoiser already operates on $k$-NN graphs over continuous coordinates. Open-ST~\cite{OpenST_Cell_2024}, with paired H\&E and subcellular ST over serial sections, is a natural 3D benchmark for this, and pretraining on HEST-1k~\cite{HEST2024} may improve cross-dataset generalization and reduce the stack-specific ST supervision needed. Further directions are retrieval beyond $z{\pm}1$ for long-range anatomy and per-spot uncertainty from repeated trajectories (\cref{supp:point-estimate}).

\section*{Acknowledgments}
We thank the Human Developmental Biology Resource (HDBR) and the Wellcome Sanger Institute Cellular Genetics IT and Spatial teams. This work was supported by Wellcome Trust institutional grants to the Wellcome Sanger Institute, by the Research Council of Finland (340273), and by the Doctoral Programme in Science, University of Helsinki.

\bibliographystyle{splncs04}
\bibliography{main}

\appendix
\clearpage
\setcounter{page}{1}

\renewcommand\thesection{S\arabic{section}}\setcounter{section}{0}
\renewcommand\thefigure{S\arabic{figure}}\setcounter{figure}{0}
\renewcommand\thetable{S\arabic{table}}\setcounter{table}{0}

% \section{Rationale}
% \label{sec:rationale}
% % 
% Having the supplementary compiled together with the main paper means that:
% % 
% \begin{itemize}
% \item The supplementary can back-reference sections of the main paper, for example, we can refer to \cref{sec:intro};
% \item The main paper can forward reference sub-sections within the supplementary explicitly (e.g. referring to a particular experiment); 
% \item When submitted to arXiv, the supplementary will already included at the end of the paper.
% \end{itemize}
% % 
% To split the supplementary pages from the main paper, you can use \href{https://support.apple.com/en-ca/guide/preview/prvw11793/mac#:~:text=Delete%20a%20page%20from%20a,or%20choose%20Edit%20%3E%20Delete).}{Preview (on macOS)}, \href{https://www.adobe.com/acrobat/how-to/delete-pages-from-pdf.html#:~:text=Choose%20%E2%80%9CTools%E2%80%9D%20%3E%20%E2%80%9COrganize,or%20pages%20from%20the%20file.}{Adobe Acrobat} (on all OSs), as well as \href{https://superuser.com/questions/517986/is-it-possible-to-delete-some-pages-of-a-pdf-document}{command line tools}.
% 

\section{Additional Method, Implementation, and Ablation Details}
\label{app:method-details}

\subsection{Flow-matching background: ZINB prior and the STFlow denoiser}
\paragraph{Zero-Inflated Negative Binomial (ZINB).} \label{pre:zinb}
A practical strength of FM is that the objective above does not depend on the form of $p_0$, so we can choose a starting prior that matches the domain to shorten the transport. \emph{STFlow} adopts a ZINB prior for gene counts due to over-dispersion and excess zeros in ST data~\cite{HuangSTFlow}. Following the scVI ~\cite{lopez2018scvi, gayoso2022scvitools}, for each gene $g$ and spot $i$ we use mean $\mu_{i,g}\!>\!0$, inverse-dispersion $\theta_g\!>\!0$, and zero-inflation probability $\pi_{i,g}\!\in\!(0,1)$, with Negative Binomial pmf
\begin{equation}
\mathrm{NB}\!\left(y\mid \mu,\theta\right)=\frac{\Gamma(y+\theta)}{\Gamma(\theta)\,y!}\!\left(\frac{\theta}{\theta+\mu}\right)^{\theta}\!\left(\frac{\mu}{\theta+\mu}\right)^y,\quad y\in\mathbb{N},
\end{equation}
and ZINB mixture
\begin{equation}
p\!\left(y\mid \mu,\theta,\pi\right)=\pi\,\mathbbm{1}\{y=0\}+(1-\pi)\,\mathrm{NB}\!\left(y\mid \mu,\theta\right).
\end{equation}
Here $y$ is the gene-count random variable, $\Gamma(\cdot)$ is the gamma function, and the ZINB parameters may include library-size factors in $\mu_{i,g}$ and logit parameterization for $\pi_{i,g}$ as in \textsc{scvi}. We sample $\mathbf{Y}_i\sim\prod_{g=1}^{G}\mathrm{ZINB}(\mu_{i,g},\theta_g,\pi_{i,g})$ and map to the model domain via $\mathbf{x}_{0,i}=\log(1+\mathbf{Y}_i)$ \label{eq:log1p}. ZINB is standard for scRNA-seq and ST likelihoods~\cite{risso2018zinbwave} and is implemented in \textsc{scvi-tools}~\cite{gayoso2022scvitools}.

\paragraph{STFlow denoiser.}
The denoiser $f_\theta$ operates on a within-section k-NN graph $G=(V,E)$ to perform spatial denoising conditioned on histology. Each node $i$ forms a token by concatenating the current state $\mathbf{x}_{t,i}$, a time embedding $e(t)$, positional features derived from $\mathbf{s}_i$, and its image embedding $\mathbf{v}_i$. Multi-head attention aggregates neighboring evidence with edge-aware biases $b_{ij}$ (functions of spatial relations such as $\|\mathbf{s}_i{-}\mathbf{s}_j\|$), using
\begin{equation}
\alpha_{ij}=\mathrm{softmax}_j\!\Big(\tfrac{\mathbf{q}_i^\top \mathbf{k}_j}{\sqrt{d}}+b_{ij}\Big),\quad
\mathbf{h}_i=\!\sum_{j\in\mathcal{N}(i)}\alpha_{ij}\,\mathbf{v}_j^{\text{(val)}},
\end{equation}
where $\mathbf{q}_i,\mathbf{k}_j,\mathbf{v}_j^{\text{(val)}}$ are the query, key, and value vectors for spots $i$ and $j$, $d$ is the per-head key/query dimensionality, $\mathcal{N}(i)$ is the k-NN set of $i$, $\alpha_{ij}$ are attention weights, and $\mathbf{h}_i$ is the aggregated representation. This graph-attention view follows established formulations~\cite{velickovic2018gat} and can be equipped with inducing-point self-attention for low-rank global mixing~\cite{lee2019settransformer}; see also the whole-slide, locally attentive denoiser in STFlow~\cite{HuangSTFlow}.

\subsection{Training algorithm}
\begin{algorithm}[t]
\small
\caption{\method training algorithm}
\label{alg:train}
\begin{algorithmic}[1]
\Require Dataset $\{(I_i,\mathbf{y}_i,\mathbf{s}_i)\}$; encoder $\phi$; kNN size $k$; ControlNet scale $\alpha(t)$; inducing count $m$; time sampler $t\sim\mathcal{T}$; optimizer (AdamW); early-stopping on validation.
\Ensure Trained denoiser $\theta$, ControlNet $\omega$; frozen ZINB prior $\psi_\eta$.

\Statex \textbf{Phase A: Pretrain ZINB prior (frozen later)}
\For{\textbf{epoch} $=1,\dots,E_{\text{ZINB}}$}
  \For{\textbf{minibatch} $\mathcal{B}$}
    \State Predict ZINB parameters $(\boldsymbol{\mu}_i,\boldsymbol{\pi}_i){=}\psi_\eta\big(\log(1{+}\mathbf{y}_i)\big)$ for $i\!\in\!\mathcal{B}$, with per-gene $\boldsymbol{\theta}$.
    \State Accumulate ZINB NLL using Eq.~\eqref{eq:zinb-nll}; update $\eta$.
  \EndFor
  \State Early-stop on validation NLL if applicable.
\EndFor
\State Freeze $\psi_\eta$ (no gradients in Phase B).

\Statex \textbf{Phase B: Flow-matching training}
\For{\textbf{step} $=1,\dots,T$}
  \State Sample a minibatch $\mathcal{B}$ of spots (possibly across sections); sample $t\!\sim\!\mathcal{T}$; compute H\&E embeddings $\mathbf{v}_i\!\leftarrow\!\phi(I_i)$.
  \State \textbf{Start sampling:} for each $i\!\in\!\mathcal{B}$, sample counts via Eq.~\eqref{eq:zinb-sample} using frozen $\psi_\eta$, then map to model space with $\mathbf{x}_{0,i}$.
  \State \textbf{Interpolant:} compute $\mathbf{x}_{t,i}$ using Eq.~\eqref{eq:fm-interp}.
  \State \textbf{Adjacent conditioning (train):}
      find $j^{\pm}(i)$ by Eq.~\eqref{eq:adj-candidates};
      adjacent tokens $\tilde{\mathbf{g}}^{\pm}_i$ by Eq.~\eqref{eq:adj-token-train}.
  \State \textbf{ControlNet:} compute $\mathbf{u}^{(b)}_i$ by Eq.~\eqref{eq:controlnet-u}; add to the input tokens via Eq.~\eqref{eq:controlnet-inject}.
  \State \textbf{Backbone with GSA:} run the spatial transformer, applying GSA read and write per Eqs.~\eqref{eq:gsa-read}--\eqref{eq:gsa-write}.
  \State \textbf{Prediction \& loss:} obtain $\hat{\mathbf{y}}_i$; backpropagate the FM loss $\mathcal{L}_\mathrm{fm}$; update $\theta,\omega$ (and any unfrozen heads).
  \State Apply gradient clipping, mixed precision, and early-stopping on validation sections as needed.
\EndFor
\State \textbf{Return} $\theta,\omega$ (trained), $\psi_\eta$ (frozen).
\end{algorithmic}
\end{algorithm}

\subsection{Implementation details}
\paragraph{Implementation details.}
All experiments use PyTorch with mixed precision, Adam optimizer~\cite{kingma2015adam}, and a single NVIDIA H100 (80\,GB). H\&E tiles are encoded by the UNI pathology foundation model~\cite{UNI_NatMed2024}. Our ZINB decoder follows the scVI parameterization and NLL~\cite{lopez2018scvi,gayoso2022scvitools}, taking as input either the STFlow latent representation or features from the frozen UNI pathology encoder~\cite{UNI_NatMed2024}. In the UNI baseline, we keep the UNI backbone fixed and train only this ZINB head, yielding an image-only encoder–decoder model without any flow-based components. Flow training uses displacement/rectified flow matching~\cite{lipman2022flowmatching,liu2023rectifiedflow}, and our ControlNet-style conditioning for cross-section guidance follows~\cite{zhang2023controlnet}. The training follows \cref{alg:train}.

\subsection{Computational complexity: detailed analysis}
This favorable scaling is a direct consequence of the design.
First, adjacent retrieval is a single nearest-neighbour query per spot on each of the two adjacent sections ($k$-d tree, $O(\log N)$ per query), negligible relative to attention.
Second, the ControlNet that injects cross-section cues is a shallow CNN/MLP whose cost is linear in the number of spots.
Third, the backbone replaces quadratic global attention with local $k$NN attention and an optional low-rank inducing-attention module that scales as $O(Nm)$ (with $m\!\ll\!N$) instead of $O(N^2)$.
Together, these choices keep both FLOPs and memory near-linear in $z$ during training and reduce the inference footprint (adjacent sections only).

\subsection{Qualitative comparison (ControlNet vs.\ STFlow)}
\cref{fig:controlnet} shows imputed expression of the erythroid gene AHSP on a held-out embryo section at flow steps 1, 3 and 5. STFlow's prediction is still speckled at step~1 and reaches only moderate values in the high-expression region by step~5, whereas \method already localizes this region at step~1 and, by steps 3--5, recovers both the high-expression cluster and the weaker neighbouring domain at intensities close to the ground truth. Neither model reproduces the small isolated hot spots near the section boundary.
\begin{figure}[!htb]
    \centering
    \includegraphics[width=0.8\linewidth]{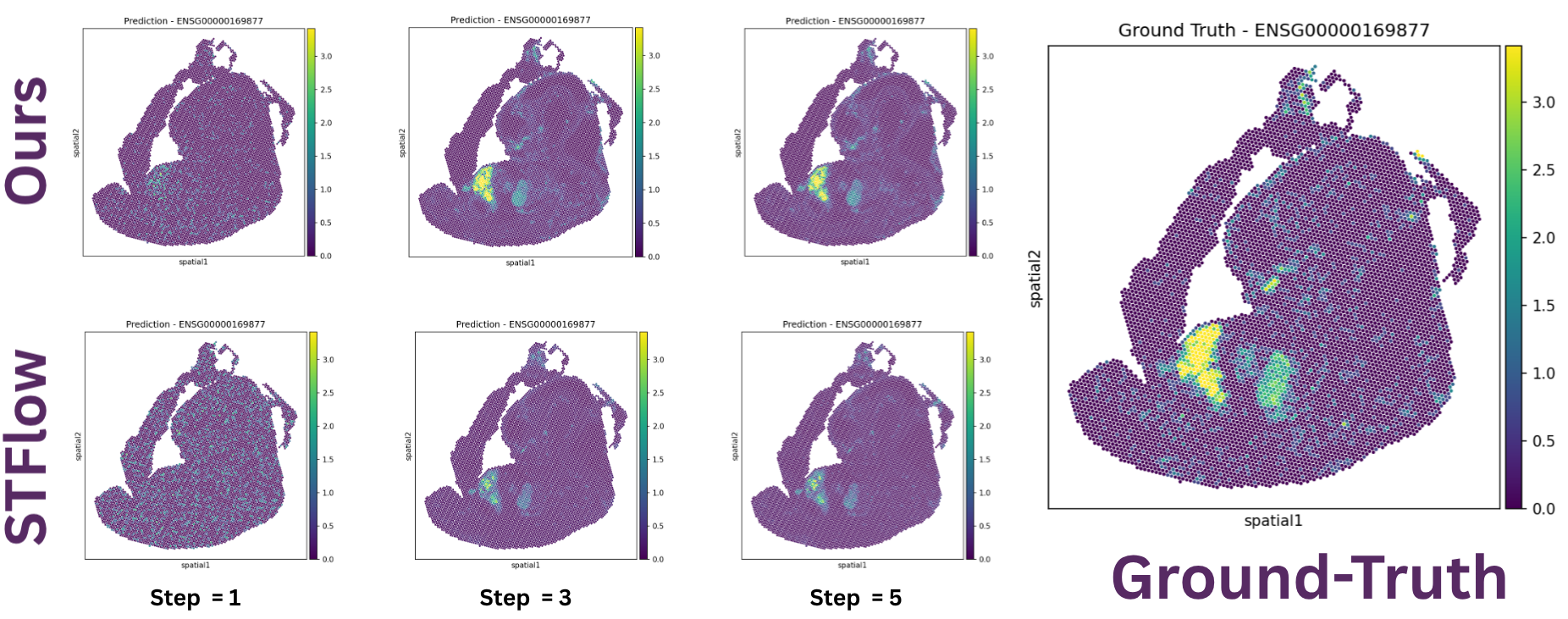}
    \caption{Imputed AHSP expression on a held-out embryo section at flow steps 1, 3 and 5: \method (top), STFlow (bottom), and ground truth (right).}
    \label{fig:controlnet}
\end{figure}

\subsection{Dataset sparsity and difficulty}
\paragraph{Data sparsity and dataset difficulty.}
% We quantify the varying degrees of sparsity and data quality across our evaluation suite in Table~\ref{tab:data_quality}. The datasets represent a broad spectrum of difficulty: {STData} serves as a high-density reference with low sparsity (13.51\%) and robust gene detection. {Embryo} reflects a typical spatial transcriptomics scenario with moderate sparsity ($\sim$69\%), while {HER2} presents a highly sparse test case with 81.12\% sparsity and fewer than 29 genes detected per spot on average. These results demonstrate that while dense datasets provide strong local signals, the high sparsity in HER2 necessitates the use of 3D contextual guidance and imputation to recover meaningful biological patterns that are otherwise lost to dropout.
The datasets represent a broad spectrum of difficulty: {STData} serves as a high-density reference with low sparsity (13.51\%) and robust gene detection. {Embryo} reflects a typical spatial transcriptomics scenario with moderate sparsity ($\sim$69\%), while {HER2} presents an extreme "stress test" case with 81.12\% sparsity and fewer than 29 genes detected per spot on average.

\section{Dataset Details}
\subsection{In-house embryo dataset.}
To generate the in-house Embryo Spatial Transcriptomics (ST) dataset, we performed \textit{10x Genomics Visium} experiments on 62 serial sagittal sections obtained from a single Formalin-Fixed Paraffin-Embedded (FFPE) tissue block of a whole human embryo at Carnegie Stage~17 and after quality control, we retained 52 slides for our benchmark. Each section encompassed the full anatomical landscape of the embryo, covering major organ systems and developing structures. The FFPE block was sectioned at a thickness of 5~$\mu$m using a \textit{Leica RM2235} microtome. Each Cytassist capture area contained an array of barcoded spots for spatially resolved mRNA capture and gene expression profiling.

The spatial transcriptomic data were processed using the \textit{Space Ranger}~v2.10 pipeline with standard quality control procedures. In total, the dataset comprises 327{,}532 spatial spots, with a median of 7{,}694 spots under tissue. 

\subsection{Public datasets.}
\paragraph{HER2 dataset.}
For the HER2 cohort we used the spatial transcriptome sequencing data from eight HER2 positive human breast tumors, where each tumor contributes three or six spatial transcriptomics sections, giving a total of 36 profiled sections. These samples were generated with the first generation Spatial Transcriptomics in situ capturing arrays and sequenced on an Illumina NextSeq 550 platform, yielding whole transcriptome counts for each capture spot. The original study combines these data with single cell RNA sequencing from additional HER2 positive tumors to deconvolve each spot into proportions of major cell types, including epithelial, myeloid, T, B, plasma, endothelial and cancer associated fibroblast populations, and reports that a typical spot aggregates roughly 0 to 200 cells. A pathologist manually annotated one section per tumor into regions such as in situ cancer, invasive cancer, immune infiltrate, adipose tissue and connective tissue, and expression based clustering plus pathway enrichment were then used to define cancer and immune core signatures and to map tertiary lymphoid like structures and type I interferon related immune niches across all 36 sections.

\paragraph{ST-data (Integrating spatial gene expression and breast tumour morphology via deep learning).}
The ST-data cohort originates from the work of He et al., who introduced ST Net using a spatial transcriptomics dataset that links histopathology to local gene expression in breast cancer. The public release contains triplicate spatial transcriptomics sections from 23 patients spanning luminal A, luminal B, HER2-positive, and triple negative subtypes, for a total of 68 sections and 30,612 spatial locations with matched hematoxylin and eosin whole slide images. Each capture spot has a diameter of about 100 micrometer on the original in situ capturing array, and the corresponding histology was digitized at 20x, so that a 224 by 224 pixel image patch centered on each spot covers roughly 150 by 150 micrometer of tissue. The sequencing depth in the original study results in a median of around 3,400 counts and several thousand detected genes per spot, and ST Net was trained to predict the log expression of a curated panel of 250 target genes from these image patches using a leave one patient out cross validation protocol. This dataset therefore, provides a well-curated benchmark with consistent pre-processing, subtype labels, and tumor or normal tissue annotations, which we reuse as a standard testbed for evaluating our method on two-dimensional sections.

\subsection{Spatial alignment of Visium slides}
\label{app:spatial-alignment}

\paragraph{Data alignment.}
For the \textbf{public} datasets (\emph{HER2} and \emph{ST-data}), we obtained the raw data from the original studies’ official repositories and did \emph{not} perform any additional alignment beyond the preprocessing carried out by the original authors. In contrast, for our \textbf{in-house embryo} dataset, we applied the two-stage alignment workflow described below.

\paragraph{Stage 1: Pre-alignment with MOSCOT.}
We first pre-aligned a total of 52 individual Visium slides non-consecutive but ordered according to their sectioning sequence using \emph{MOSCOT}\cite{Klein2025moscot}. MOSCOT leverages spatial transcriptomics to align sets of two dimensional Visium spots into a coherent pseudo three dimensional volume. All sections were registered to a single reference slide selected from the middle of the stack, and the corresponding transformation matrices produced by MOSCOT were extracted. These matrices were then applied to pre-align the full-resolution H\&E images, which we downscaled by a factor of two, resulting in a final pixel size of approximately $0.45\,\mu$m.

\paragraph{Stage 2: Fine registration with \texttt{pystackreg}.}
The pre-aligned image stack served as input for fine-grained image registration implemented with the Python package \emph{pystackreg}. We used the green (G) channel of the H\&E images providing the highest contrast for alignment. Affine registration (translation, rotation, scaling, shear) was performed sequentially: each section was registered to its immediate predecessor (e.g., section 2 to section 1, section 3 to section 2, and so on). This produced a finely registered image stack and an updated set of transformation matrices. 

\paragraph{Final spot-coordinate update.}
To obtain Visium spot coordinates corresponding to the fully registered images, we applied the transformation matrices from the fine-registration step to the MOSCOT-aligned spot positions, yielding the final coordinates used in downstream analyses. The implementation will be released following peer review.

\paragraph{Use of held-out sections in registration.}
MOSCOT pre-alignment uses the ST of all 52 sections, including those held out for evaluation, but only to estimate rigid/affine transformations of spot coordinates and H\&E images; no expression values are transferred, averaged or propagated between sections. Registration therefore determines \emph{where} a held-out spot sits in the common 3D frame, not \emph{what} it is predicted to express, and held-out expression is never used for training, prior construction, adjacent-section conditioning or inference. Only the MOSCOT stage uses ST; the fine registration with \texttt{pystackreg} uses H\&E alone. Replacing MOSCOT with an image-only pipeline such as VALIS~\cite{Gatenbee2023VALIS} removes ST from registration entirely; we plan to use it in future studies.

\paragraph{Registration quality.}
We check the registration with expression: for a pair of consecutive registered sections of the in-house embryo ($\Delta z{=}70\,\mu$m), we shift one section by $(dx,dy)\in[-200,200]^2\,\mu$m, match every spot to its nearest spot on the other section as \method does, and record the mean per-spot PCC of the matched expression. The best shift, an estimate of the residual misalignment, is $22\,\mu$m, well below the Visium spot pitch of $100\,\mu$m, and the matched-spot correlation without shift is nearly the same as at the best shift ($0.750$ vs.\ $0.751$).

\subsection{Gene panel definitions.}
For completeness, we summarize the three gene panels used throughout the main paper.
\begin{itemize}
  \item \textbf{HVG panel.} A highly variable gene (HVG) panel obtained per dataset using a standard Scanpy pipeline (library-size normalization, $\log(1 + x)$ transform, and variance ranking). In the main paper we report results for the HVG-250 configuration.
  \item \textbf{Custom panel.} About 150 genes that are highly variable across the whole embryo stack: per-section HVGs (Scanpy, Seurat v3, top 2{,}000) are grouped into 11 bins of 4--5 sections spread along $z$; within each bin the 50 genes shared by most sections are kept, and the bins are unioned.
  \item \textbf{Marker panel.} A curated set of $211$ marker genes associated with $106$ anatomical regions in the embryo stack. This panel is used for the biological-validity analyses (\cref{sec:biological-validity}).
\end{itemize}
These panels correspond to the sets referred to as (i) HVG, (ii) custom, and (iii) marker genes in the main text.

\section{Experiments Details}
\label{app:experiments_details}
\noindent\textbf{Spot-centric visual features.} For each spatial spot, we extract a $224{\times}224$ image patch centered at the spot from the whole-slide image, pass it through a frozen pre-trained foundation encoder, and use the resulting global embedding as the node feature for that spot.

The key hyperparameters used to train the model are summarized in the table \ref{tab:s-core-hparams}.

\noindent\textbf{Dataset splits.} For all public datasets, we follow the official train/test partitions used in the ASIGN paper~\cite{Zhu_2025_CVPR_ASIGN} to ensure fair comparisons. Unless otherwise noted, all metrics are reported on the held-out  sections.

\paragraph{Baseline training protocol.}
For all baselines built on pathology foundation encoders (UNI, CIGA, and GigaPath) we follow a consistent training protocol that mirrors HoloTea as closely as possible. The pathology encoder weights are kept frozen, and we train only the downstream ZINB decoder and, where applicable, the flow-matching components, using the optimization hyperparameters listed in Table~S3. This ensures that differences in performance primarily reflect the choice of encoder and our 3D-aware architecture rather than differences in training budget. For the ASIGN (3D) baseline we adapt the official configuration to our preprocessing and normalization pipeline and train until validation performance plateaus.

\noindent\textbf{Alternative pathology encoders.}
In addition to the UNI pathology backbone in the main experiments, we also study the
effect of swapping the image encoder while keeping the rest of the pipeline fixed.
Specifically, in the appendix we repeat our experiments with two frozen H\&E encoders:
(i) the self-supervised contrastive histopathology model of Ciga et al.~\cite{ciga2022self},
pretrained on large collections of unlabeled whole-slide images; and
(ii) the GigaPath whole-slide foundation model for digital pathology~\cite{Xu2024GigaPath}.
For both variants, we extract patch features with the pretrained encoder, freeze all
its weights, and train only the ZINB decoder (and, where applicable, the flow-matching
components) using the same optimizer, learning rate schedule, and training budget as
in the UNI setup. This yields image-only encoder–decoder baselines that isolate the
impact of the pathology backbone on 3D imputation performance.

\paragraph{Global context tokens $m$.}
All experiments use the global context mechanism described in the main paper with $m{=}64$ inducing tokens for every dataset. Tables~S6--S8 report a sensitivity study over $m\in\{0,4,8,\dots\}$; the $m{=}0$ row is STFlow. In practice, moderate values of $m$ already capture most of the benefit, while very large values provide only marginal additional gains at a higher computational cost.

\begin{table}[t]
\centering
\caption{\textbf{Core model hyperparameters (defaults).}}
\label{tab:s-core-hparams}
\begin{tabular}{@{} l p{0.3\columnwidth} @{}}
\toprule
\textbf{Parameter} & \textbf{Default} \\
\midrule
Backbone layers $L$ & 4 \\
Hidden dimension $d$ & 128 \\
Pairwise / edge dimension & 128 \\
Attention heads & 4 \\
$k$ nearest neighbors & 8 \\
Dropout / Attention dropout & 0.2 / 0.2 \\
Global context (inducing tokens $m$) & $64$ \\
\bottomrule
\end{tabular}
\end{table}

\begin{table}[h]
\centering
\caption{\textbf{Training and inference hyperparameters.}}
\label{tab:s-train-hparams}
\begin{tabular}{ll}
\toprule
\textbf{Parameter} & \textbf{Default / Setting} \\
\midrule
Batch size (slide-level) & 2 \\
Epochs & 100 \\
Optimizer / Learning rate & Adam / $5\times10^{-4}$ \\
Gradient clipping (L2) & 1.0 \\
Time conditioning & Sinusoidal $\rightarrow$ MLP \\
Inference steps ($S$) & 5 \\
\bottomrule
\end{tabular}
\end{table}

\begin{table}[t]
\centering
\caption{\textbf{Priors and optional modules.}}
\label{tab:s-prior-hparams}
\begin{tabular}{@{} p{0.3\columnwidth} p{0.6\columnwidth} @{}}
\toprule
\textbf{Component} & \textbf{Default / Setting} \\
\midrule
Prior sampler & ZINB (default) \\
ZINB parameters & total\_count $= 1$, logits $= 0.1$, zi\_logits $= 0.0$ \\
Learned ZINB prior ($\phi_e$) & optional (frozen at train time) \\
ControlNet conditioning & optional; channels $= 32$, grid $= 64$, scale $= 1.0$ \\
\bottomrule
\end{tabular}
\end{table}

\begin{table}[h]
\centering
\caption{\textbf{Foundation encoders and node-feature dimensions.} Encoders are frozen.}
\label{tab:s-encoders}
\begin{tabular}{lll}
\toprule
\textbf{Encoder} & \textbf{Patch size} & \textbf{Output dim} \\
\midrule
UNI v1 (official) & $224{\times}224$ & 1024 \\
GigaPath & $224{\times}224$ & 1536 \\
CIGA & $224{\times}224$ & 512 \\
\bottomrule
\end{tabular}
\end{table}

\section{Results Details}
\label{app:results_details}

\paragraph{Complexity and dataset scale.}
The embryo dataset used in our experiments consists of $327{,}532$ spatial spots across the 3D stack (see Table~S9). The complexity numbers reported in \cref{tab:complexity_spots_flops_mem} are measured per slide within this stack and should be interpreted in the context of this overall scale: the global attention backbone with $m$ inducing tokens enables training and inference to scale approximately linearly with the number of spots per slide and with the depth of the stack.

\paragraph{Randomness and variability.}
Unless otherwise stated, all metrics in the main paper are reported for a single trained model per configuration. For each metric we aggregate over all spots or all genes in the corresponding test set, which substantially reduces statistical noise. Additional runs with different random seeds on a subset of configurations showed only minor fluctuations in the reported metrics, so we omit per-seed averages for brevity and focus on the main configuration in the tables.

\paragraph{Point estimate and the role of the generative formulation.}
\label{supp:point-estimate}
The reported imputation is the last-step endpoint prediction $\hat{\mathbf{y}}_i$ of a single trajectory. Because $f_\theta$ is trained with the squared-error loss $\mathcal{L}_\mathrm{fm}$, $\hat{\mathbf{y}}_i$ estimates the conditional mean of $\mathbf{y}_i$ given the current state and context, so the intended output is a best point estimate rather than one plausible sample. The generative formulation serves to model all spots of a section jointly, to start from a count-aware, cross-section-informed distribution, and to condition each spot on the measured expression of the matched spots on the adjacent profiled sections. Repeated trajectories from different start samples could quantify per-spot uncertainty, which we leave to future work.

% \subsection{Extended expression grid}

\begin{figure*}[t]
  \centering
  \includegraphics[width=\linewidth]{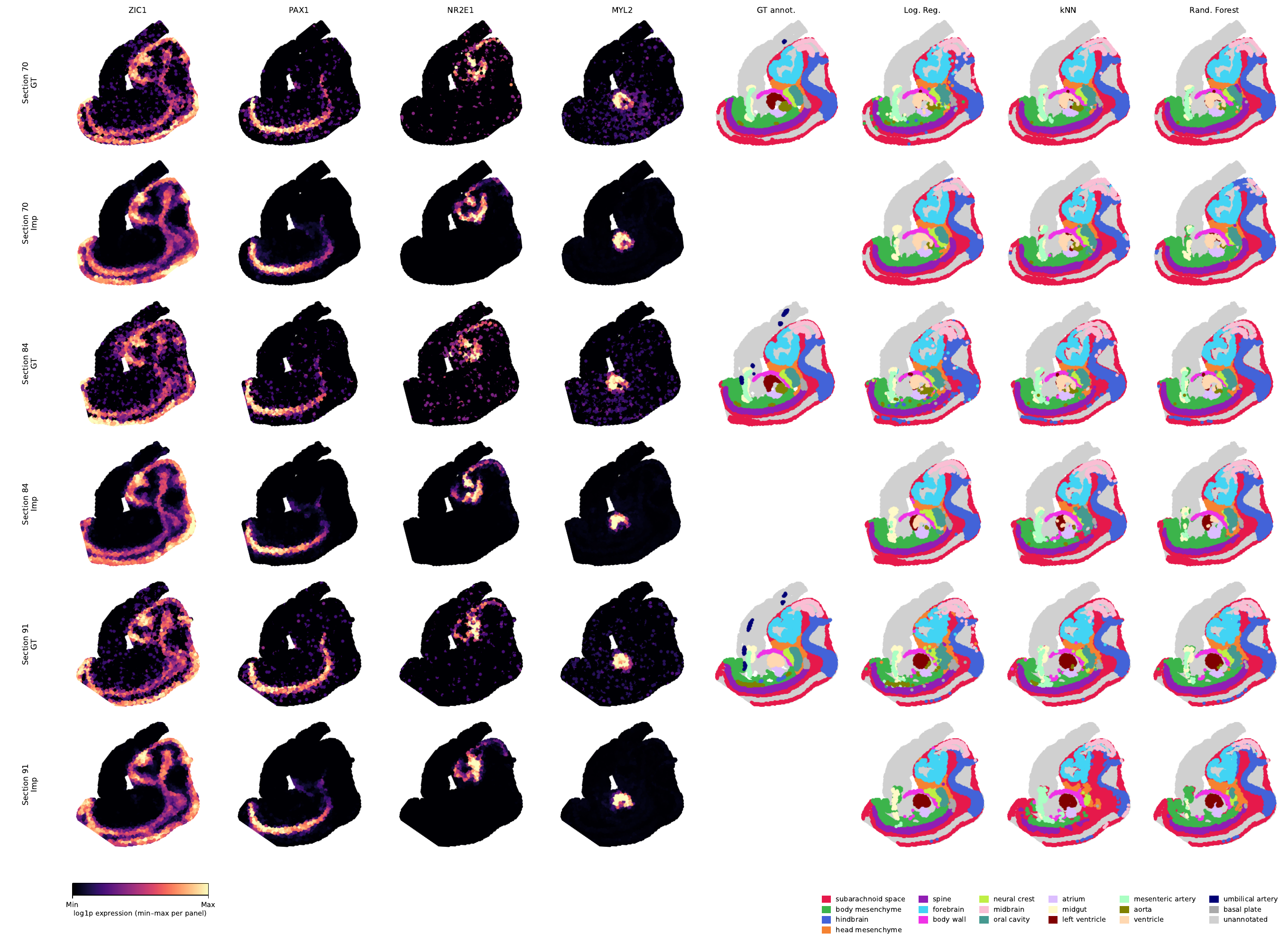}
  \caption{%
    \textbf{Extended spatial gene expression grid across all three annotated sections.}
    Each pair of rows shows GT and \method imputed expression for sections 70, 84, and 91.
    Gene expression columns (ZIC1, PAX1, NR2E1, MYL2) are followed by four annotation columns: GT anatomical annotation (GT rows only); logistic regression, $k$-nearest neighbour ($k{=}15$), and random forest predictions trained on GT or imputed expression.
    \label{suppl-fig:expression-grid}
  }
\end{figure*}

% \subsection{Per-region anatomy classification accuracy}

\begin{figure*}[t]
  \centering
  \includegraphics[width=0.75\linewidth]{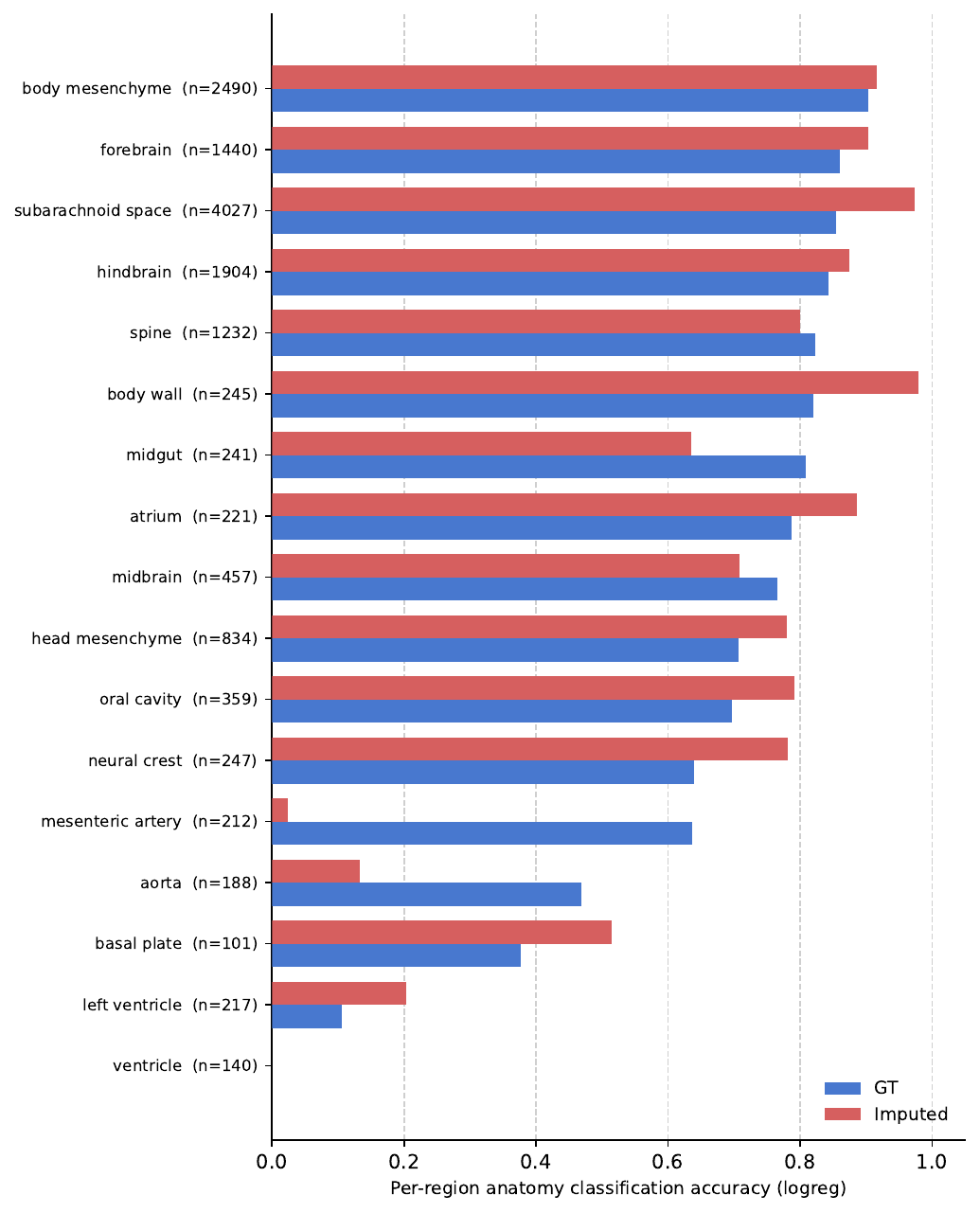}
  \caption{%
    \textbf{Per-region anatomy classification accuracy (logistic regression, leave-one-section-out CV).}
    Each bar pair shows accuracy for one of the 19 annotated anatomical regions, comparing classifiers trained on GT (blue) vs.\ imputed (red) expression of the 211-gene panel across sections 70, 84, and 91.
    Regions are sorted by GT accuracy.
    \label{suppl-fig:region-accuracy}
  }
\end{figure*}

% \subsection{Anatomy classifier accuracy and F1 scores}

\begin{table}[t]
  \centering
  \caption{%
    \textbf{Anatomy classification performance (leave-one-section-out CV).}
    Accuracy and macro-averaged F1 score for classifiers trained on GT vs.\ \method imputed expression of the 211-gene panel (sections 70, 84, 91; 19 anatomical classes).
    Bold indicates the higher value within each GT/Imp pair.
    \label{suppl-tab:classifier-metrics}
  }
  \small
  \begin{tabular}{@{} ll rr rr @{}}
    \toprule
    & & \multicolumn{2}{c}{\textbf{Accuracy}} & \multicolumn{2}{c}{\textbf{Macro F1}} \\
    \cmidrule(lr){3-4} \cmidrule(lr){5-6}
    \textbf{Classifier} & \textbf{Section} & \textbf{GT} & \textbf{Imp} & \textbf{GT} & \textbf{Imp} \\
    \midrule
    \multirow{4}{*}{Logistic reg.}
      & 70      & \textbf{0.829} & 0.824          & \textbf{0.696} & 0.608 \\
      & 84      & 0.773          & \textbf{0.882} & 0.633          & \textbf{0.701} \\
      & 91      & 0.820          & \textbf{0.834} & \textbf{0.659} & 0.628 \\
      & Overall & 0.807          & \textbf{0.847} & \textbf{0.664} & 0.657 \\
    \midrule
    \multirow{4}{*}{$k$-NN ($k{=}15$)}
      & 70      & \textbf{0.866} & 0.800          & \textbf{0.745} & 0.653 \\
      & 84      & 0.742          & \textbf{0.787} & 0.607          & \textbf{0.638} \\
      & 91      & \textbf{0.828} & 0.557          & \textbf{0.649} & 0.374 \\
      & Overall & \textbf{0.812} & 0.716          & \textbf{0.672} & 0.577 \\
    \midrule
    \multirow{4}{*}{Random forest}
      & 70      & \textbf{0.847} & 0.809          & \textbf{0.649} & 0.569 \\
      & 84      & 0.768          & \textbf{0.811} & 0.588          & \textbf{0.623} \\
      & 91      & \textbf{0.821} & 0.725          & \textbf{0.610} & 0.450 \\
      & Overall & \textbf{0.812} & 0.782          & \textbf{0.618} & 0.561 \\
    \bottomrule
  \end{tabular}
\end{table}

% \subsection{Per-region marker gene enrichment}

\begin{figure*}[t]
  \centering
  \includegraphics[width=\linewidth]{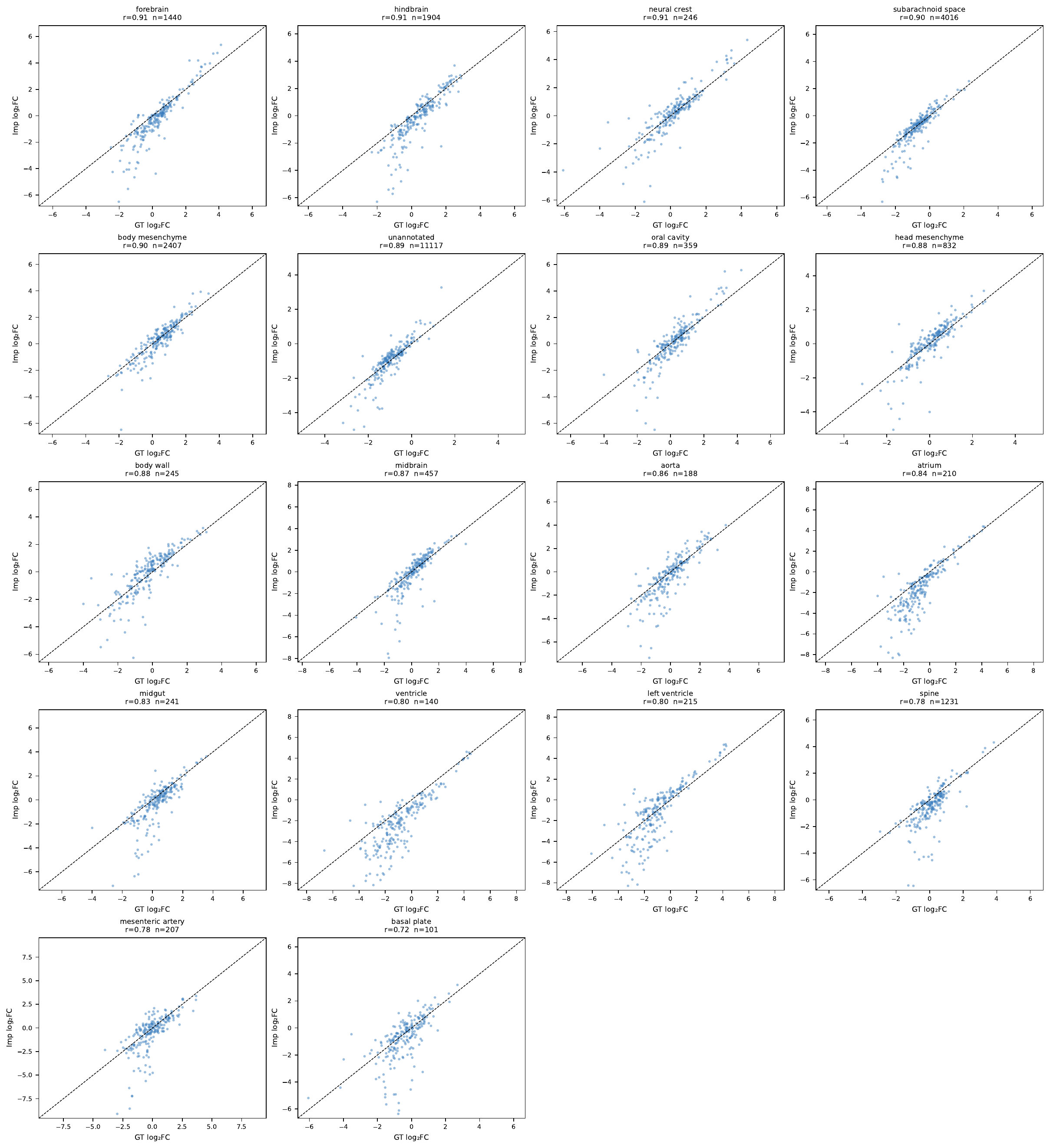}
  \caption{%
    \textbf{Per-region marker gene enrichment correlation (GT vs.\ imputed).}
    Each panel shows the per-gene log2 fold-change between one annotated anatomical region and the rest, computed for GT (x-axis) and \method imputed expression (y-axis) across sections 70, 84, and 91.
    Spearman $r$ is shown per panel.
    \label{suppl-fig:enrichment}
  }
\end{figure*}

% \subsection{Pathway score scatter plots}

\begin{figure*}[t]
  \centering
  \includegraphics[width=\linewidth]{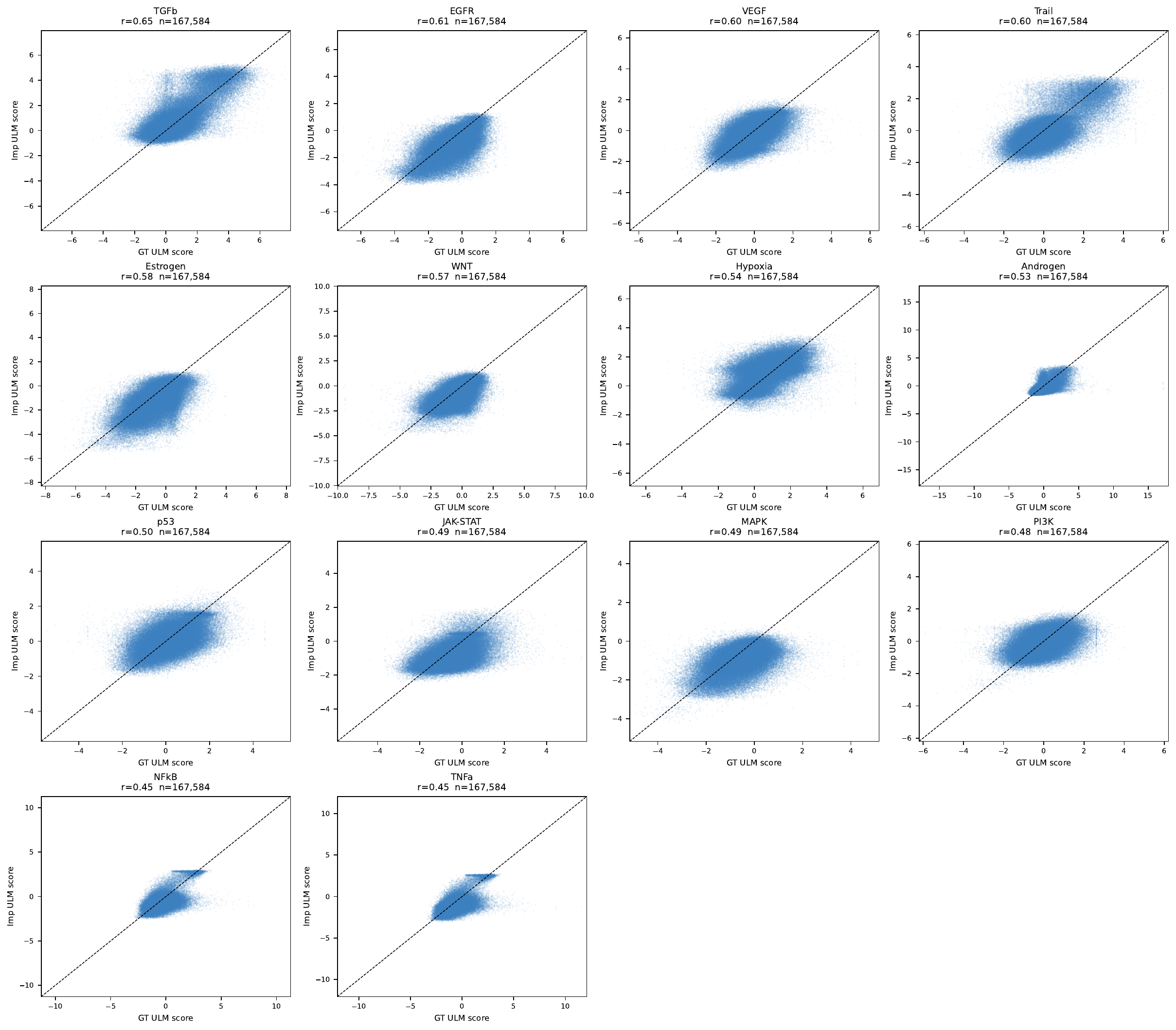}
  \caption{%
    \textbf{Per-pathway PROGENy activity score scatter (GT vs.\ imputed).}
    Each panel shows per-spot GT (x-axis) vs.\ \method imputed (y-axis) pathway activity scores for one of the 14 PROGENy signaling pathways, across all 26 held-out sections ($n{=}167{,}584$ spots).
    Spearman $r$ and $p$-value are shown per panel.
    \label{suppl-fig:pathway-scatter}
  }
\end{figure*}
% \paragraph{Clustering}
% Clustering of true and predicted counts was performed using routines implemented in the \texttt{scanpy} package.
% First, counts were normalized to \num{10000} counts per cell, and then log1p-transformed. 
% The transformed counts were projected to the top-50 principal components and 15-nearest neighbors were determined.
% Finally, clustering analysis was performed using the Leiden community detection algorithm with a resolution parameter of 1.

% \paragraph{Neighbor consistency}
% Neighbor consistency \(NC\) for the spot-neighbor graph was calculated according to
% \begin{equation}
% NC = \sum_{(v, w) \in E} \frac{\mathbbm{1}_{c(v) = c(w)}}{\left| E \right|},
% \end{equation}
% where \(E\) denotes the set of neighboring vertex pairs and \(c(v)\) indicates the cluster assignment of vertex \(v\).

% ============================ Effect of m ============================
\paragraph{Effect of \(m\) (EMBRYO).}
We sweep \(m \in \{0,4,8,16,32,64,128,256\}\) on the in-house embryo dataset hvg gene set and report MAE, MSE, \textbf{PCC SPOT} (mean\(\pm\)std across spots), and \textbf{PCC GENE} (mean\(\pm\)std across genes); see \Cref{tab:m_sweep_embryo_pow2}.
Overall, \(m\) has a clear impact: relative to the \(m{=}0\) baseline, errors decrease consistently as \(m\) increases.
In particular, MSE improves from \(0.255\) to \(\mathbf{0.202}\) (absolute change \(-0.053\), \(\sim\!20.8\%\) reduction) and MAE from \(0.270\) to \(\mathbf{0.232}\) (absolute change \(-0.036\), \(\sim\!13.3\%\) reduction) at \(m{=}256\).
Correlations also benefit: the best \textbf{PCC SPOT} is reached at a moderate \(m{=}16\) (\(\mathbf{0.572}\pm0.246\), +0.015 over baseline), while \textbf{PCC GENE} attains its highest values at both small and medium \(m\) (\(m{=}4\) and \(m{=}128\): \(\mathbf{0.681}\pm\{0.171,0.173\}\), +0.018 over baseline).

Interestingly, small values of \(m\) already capture most of the gain.
At \(m{=}4\), we obtain MAE \(0.237\), MSE \(0.207\), \textbf{PCC SPOT} \(0.571\pm0.244\), and \textbf{PCC GENE} \(\mathbf{0.681}\pm0.171\), recovering a large fraction of the improvement over \(m{=}0\).
For larger \(m \in [32,256]\), we observe diminishing returns: errors continue to decrease slightly (e.g., MSE \(0.206 \to 0.202\)), while correlations remain effectively stable within \(\sim\!0.002\!-\!0.005\).
Taken together, these results indicate that \(m\) controls a useful capacity/aggregation trade-off, but relatively small context sizes (e.g., \(m{=}4\!-\!16\)) already offer a strong balance between accuracy and efficiency.

\paragraph{Effect of \(m\) (ST).}
On the \textbf{ST} dataset (\Cref{tab:st_m_sweep_final_ST}), increasing \(m\) yields clear accuracy gains up to a moderate context.
Relative to \(m{=}0\) (MSE \(0.443\), MAE \(0.532\), \textbf{PCC SPOT} \(0.775\pm0.080\), \textbf{PCC GENE} \(0.540\pm0.142\)), performance improves steadily and peaks around \(m{=}32\):
MSE reaches \(\mathbf{0.382}\) (\(-0.061\), \(\sim\!13.8\%\) reduction), MAE \(\mathbf{0.479}\) (\(-0.053\), \(\sim\!10.0\%\) reduction), while correlations also increase to \(\mathbf{0.786}\) (SPOT; \(+0.011\)) and \(\mathbf{0.572}\) (GENE; \(+0.032\)).
Notably, most of these gains are already realized by \(m{=}8\!-\!16\) (e.g., MSE \(0.392\), MAE \(0.480\) at \(m{=}16\)), after which returns diminish and performance slightly plateaus (\(m{\ge}32\)).

\paragraph{Effect of \(m\) (HER2).}
For \textbf{HER2} , the effect of \(m\) is modest.
Compared to \(m{=}0\) (MSE \(0.477\), MAE \(0.534\), \textbf{PCC SPOT} \(0.740\pm0.125\), \textbf{PCC GENE} \(0.620\pm0.146\)), the best results occur at \(m{=}16\): MSE \(\mathbf{0.462}\) (\(-0.015\), \(\sim\!3.1\%\) reduction), MAE \(\mathbf{0.530}\) (\(-0.004\), \(\sim\!0.7\%\) reduction), \textbf{PCC SPOT} \(\mathbf{0.741}\) (\(+0.001\)) and \textbf{PCC GENE} \(\mathbf{0.623}\) (\(+0.003\)) \Cref{tab:st_m_sweep_final}. 
Beyond \(m{\approx}16\), curves remain essentially flat.
We \emph{speculate} this limited sensitivity arises from dataset and number of spot size: as summarized in \Cref{avg_num_spot}, HER2 contains relatively few spots, so enlarging \(m\) adds little additional neighborhood context.
In regimes with sparse spatial neighborhoods, the proposed module may thus offer less scope for improvement, consistent with the small but nonzero gains observed here.

% Preamble: \usepackage{booktabs,xcolor}
\newcommand{\NA}{\textemdash}

% ============ Dataset A ============
% Requires: \usepackage{booktabs}
\begin{table}[t]
  \centering
  \caption{Effect of \(m\) (powers of two) on performance for the \textbf{EMBRYO} dataset.
  This is a sensitivity (scaling) study of how much global-attention capacity helps; the \(m{=}0\) row is STFlow.
  Lower is better for MAE/MSE; higher is better for PCC SPOT / PCC GENE.}
  \label{tab:m_sweep_embryo_pow2}
  \scriptsize
  \begin{tabular}{@{} r r r p{0.3\columnwidth} p{0.3\columnwidth} @{}}
    \toprule
    \textbf{\(m\)} &
    \textbf{MSE $\downarrow$} &
    \textbf{MAE $\downarrow$} &
    \textbf{PCC SPOT (mean $\pm$ std) $\uparrow$} &
    \textbf{PCC GENE (mean $\pm$ std) $\uparrow$} \\
    \midrule
    0   & 0.255 & 0.270 & 0.557 $\pm$ 0.248 & 0.663 $\pm$ 0.166 \\
    4   & 0.207 & 0.237 & 0.571 $\pm$ 0.244 & \textbf{0.681} $\pm$ 0.171 \\
    8   & 0.212 & 0.241 & 0.570 $\pm$ 0.246 & 0.680 $\pm$ 0.173 \\
    16  & 0.210 & 0.240 & \textbf{0.572} $\pm$ 0.246 & 0.675 $\pm$ 0.174 \\
    32  & 0.206 & 0.236 & 0.567 $\pm$ 0.244 & 0.680 $\pm$ 0.173 \\
    64  & 0.204 & 0.236 & 0.568 $\pm$ 0.245 & 0.679 $\pm$ 0.170 \\
    128 & 0.203 & 0.235 & 0.569 $\pm$ 0.242 & \textbf{0.681} $\pm$ 0.173 \\
    256 & \textbf{0.202} & \textbf{0.234} & 0.569 $\pm$ 0.243 & 0.679 $\pm$ 0.177 \\
    \bottomrule
  \end{tabular}
\end{table}

% ============ Dataset B ============
% Requires: \usepackage{booktabs}

\begin{table}[t]
  \centering
  \caption{Effect of \(m\) on performance for the \textbf{ST} dataset. This is a sensitivity (scaling) study of how much global-attention capacity helps; the \(m{=}0\) row is STFlow. \cref{tab:three_datasets_metrics} uses a fixed \(m{=}64\). Lower is better for MAE/MSE; higher is better for PCC SPOT / PCC GENE.}
  \label{tab:st_m_sweep_final_ST}
  \scriptsize
  \begin{tabular}{@{} r r r p{0.3\columnwidth} p{0.3\columnwidth} @{}}
    \toprule
    \textbf{\(m\)} &
    \textbf{MSE $\downarrow$} &
    \textbf{MAE $\downarrow$} &
    \textbf{PCC SPOT (mean $\pm$ std) $\uparrow$} &
    \textbf{PCC GENE (mean $\pm$ std) $\uparrow$} \\
    \midrule
    0  & 0.443 & 0.532 & 0.775 $\pm$ 0.080 & 0.540 $\pm$ 0.142 \\
    4  & 0.412 & 0.502 & 0.781 $\pm$ 0.014 & 0.561 $\pm$ 0.101 \\
    8  & 0.403 & 0.491 & 0.782 $\pm$ 0.012 & 0.563 $\pm$ 0.098 \\
    16 & 0.392 & 0.480 & 0.784 $\pm$ 0.012 & 0.571 $\pm$ 0.012 \\
    32 & \textbf{0.382} & \textbf{0.479} & \textbf{0.786} $\pm$ 0.012 & \textbf{0.572} $\pm$ 0.038 \\
    64 & 0.384 & 0.481 & 0.780 $\pm$ 0.045 & 0.569 $\pm$ 0.019 \\
    \bottomrule
  \end{tabular}
\end{table}

% ============ Dataset C ============
\begin{table}[t]
  \centering
  \caption{Effect of \(m\) on performance for the \textbf{HER2} dataset. This is a sensitivity (scaling) study of how much global-attention capacity helps; the \(m{=}0\) row is STFlow. Lower is better for MAE/MSE; higher is better for PCC SPOT / PCC GENE.}
  \label{tab:st_m_sweep_final}
  \scriptsize
  \begin{tabular}{@{} r r r p{0.3\columnwidth} p{0.3\columnwidth} @{}}
    \toprule
    \textbf{\(m\)} &
    \textbf{MSE $\downarrow$} &
    \textbf{MAE $\downarrow$} &
    \textbf{PCC SPOT (mean $\pm$ std) $\uparrow$} &
    \textbf{PCC GENE (mean $\pm$ std) $\uparrow$} \\
    \midrule
0  & 0.477 & 0.534 & 0.740 $\pm$ 0.125 & 0.620 $\pm$ 0.146 \\
4  & 0.473 & 0.533 & 0.739 $\pm$ 0.010 & 0.618 $\pm$ 0.013 \\
8  & 0.471 & 0.533 & 0.740 $\pm$ 0.011 & 0.62 $\pm$ 0.037 \\
16 & \textbf{0.462} & \textbf{0.530} & \textbf{0.741} $\pm$ 0.023 & \textbf{0.623} $\pm$ 0.012 \\
32 & 0.472 & 0.533 & 0.740 $\pm$ 0.021 & 0.621 $\pm$ 0.023 \\
    \bottomrule
  \end{tabular}
\end{table}

% Requires: \usepackage{booktabs}
\begin{table}[t]
\label{avg_num_spot}
  \centering
  \caption{Average number of spots per dataset.}
  \label{tab:avg_spots}
  \scriptsize
  \begin{tabular}{@{} r r r @{}} 
    \toprule
    \textbf{HER2} & \textbf{ST} & \textbf{EMBRYO} \\
    \midrule
    378 & 482 & 6299 \\
    \bottomrule
  \end{tabular}
\end{table}

\begin{figure*}
  \includegraphics[width=\textwidth]{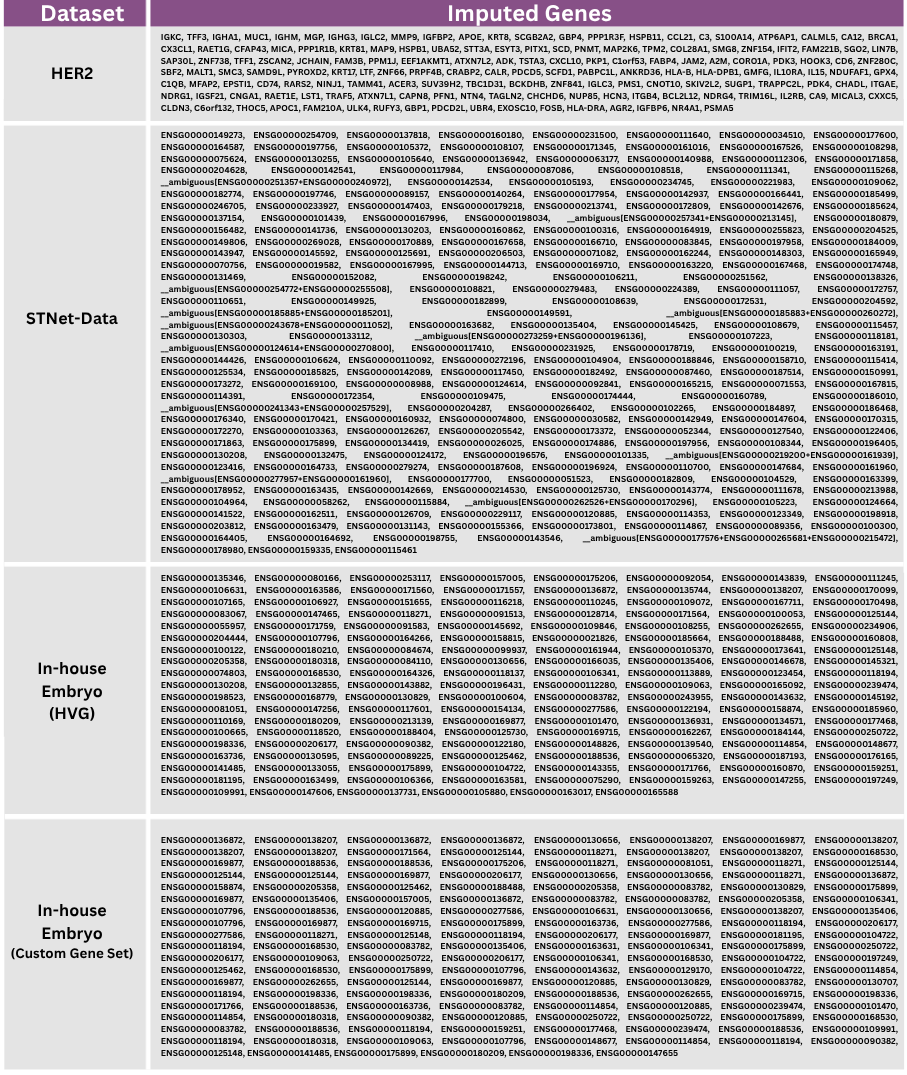}
  \caption{
    \label{suppl-fig:gene-list}
    Imputed gene panels per datasets HER2, ST-Dataset, In-house Embryo HVG and Custom
}
\end{figure*}

\section{Biological Validity of Imputed Expression}
\label{app:biological-validity}

% \subsection{Analysis methods}
% \label{app:bio-methods}

\paragraph{Anatomical label assignment.}
Each annotated spot carries a soft membership score in $[0,1]$ for each of 234 anatomical regions organised at five hierarchical levels.
For all classification and enrichment analyses we derive a single hard label per spot by taking the argmax over the level-1 region scores; spots whose maximum score equals zero are treated as unannotated.
Regions represented by fewer than 100 spots across the analysis set are excluded, yielding 19 usable classes for sections 70, 84, and 91.

\paragraph{Anatomy classifiers.}
Classifiers use the raw $\log(1{+}x)$ expression of the 211-gene panel as features without further normalisation or dimensionality reduction.
Three classifiers are evaluated: \emph{logistic regression} (L2 penalty, $C{=}1$, \texttt{lbfgs} solver, max.\ 1000 iterations); \emph{$k$-nearest neighbours} ($k{=}15$, Euclidean distance); and \emph{random forest} (200 trees).
Leave-one-section-out cross-validation trains on two sections and evaluates on the third; classifiers are always trained on GT expression and then evaluated on both GT and imputed expression of the held-out section.
Overall accuracy and macro-averaged F1 are computed by pooling predictions across all three test folds.

\paragraph{Per-region enrichment correlation.}
Region-specific enrichment is quantified as the per-gene $\log_2$ fold-change of mean expression inside a region versus all other spots (pseudocount $10^{-3}$), computed separately for GT and imputed expression.
Spots with all annotation scores equal to zero are retained as an additional ``unannotated'' group.
Spearman $r$ between the GT and imputed fold-change vectors (one value per gene, 211 in total) measures how faithfully the imputed transcriptional signature of each region mirrors the ground truth.

\paragraph{Joint UMAP embedding.}
GT and imputed expression matrices from sections 70, 84, and 91 ($14{,}571$ spots per source) are vertically concatenated into a single $29{,}142{\times}211$ matrix.
PCA is applied with 50 components (86\% cumulative variance explained), followed by $k$-NN graph construction ($k{=}15$) and UMAP projection ($\mathrm{min\_dist}{=}0.3$); all steps use Scanpy with \texttt{random\_state\,=\,0}.
Anatomy labels in panel~(d) of \cref{fig:biological-validity} are shown for GT spots only.

\paragraph{PROGENy pathway scoring.}
The full PROGENy gene--pathway weight model is retrieved from the OmniPath REST API and applied via the univariate linear model (ULM) method in decoupleR~\cite{Badia2022}, retaining only pathways with at least five target genes covered by the 211-gene panel (yielding 14 pathways with 132--207 covered genes each).
All 26 held-out sections are used; 686 spots with all-zero GT expression are excluded prior to scoring.
Activity scores are computed independently for GT and imputed expression, and Spearman $r$ is reported per pathway across all $n{=}167{,}584$ remaining spots.

\section{Extended Results}
\label{app:extended-results}

This appendix collects results referenced from the main text: the serial-stack setting (\cref{fig:teaser}), embryo custom-panel performance (\cref{tab:embryo_custom_panels}), per-slide computational complexity (\cref{tab:complexity_spots_flops_mem}), and the full component ablation (\cref{tab:ablation_component}) with the per-panel analysis below.

\begin{figure}[t]
    \centering
    \includegraphics[width=0.55\linewidth]{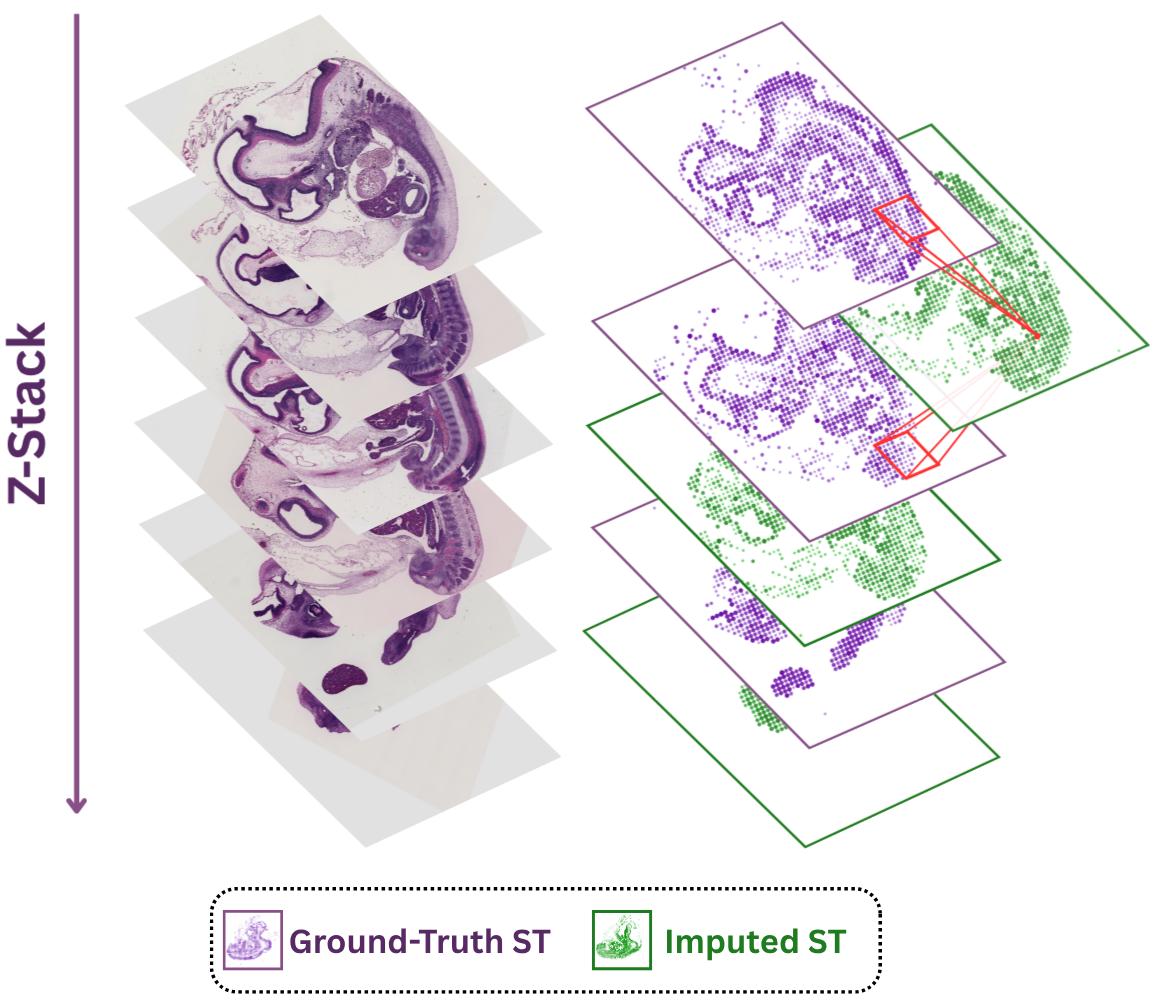}
    \caption{\method leverages spatial neighborhood information from adjacent sections to improve 3D consistency in H\&E to ST imputation.}
    \label{fig:teaser}
\end{figure}

% Requires siunitx in the preamble:
% \usepackage{siunitx}

\begin{table}[t] % single column
  \centering
  \scriptsize
  \setlength{\tabcolsep}{5pt}
  \renewcommand{\arraystretch}{1.15}
  \caption{\textbf{Embryo imputation with the custom gene panel.}
    Same split as \cref{tab:three_datasets_metrics} (split (ii), even-slice hold-out); this is not a second split. Only the gene panel differs: the custom panel ($\sim$150 genes highly variable across bins of sections spread along the stack; see \emph{Gene sets} in the experimental settings) replaces HVG-250.
    Metrics, uncertainties (standard deviation across spots for PCC Spot and across genes for PCC Gene) and model configurations are as in \cref{tab:three_datasets_metrics}; Ours is the full \method. Best results in bold.}
  \label{tab:embryo_custom_panels}
  \scalebox{0.9}{
  \begin{tabular}{
    l
    S[table-format=0.4]
    S[table-format=0.4]
    S[table-format=0.4(4), separate-uncertainty, table-align-uncertainty]
    S[table-format=0.4(4), separate-uncertainty, table-align-uncertainty]
  }
    \toprule
    \textbf{Model}
      & \multicolumn{1}{c}{\textbf{MSE} (↓)}
      & \multicolumn{1}{c}{\textbf{MAE} (↓)}
      & \multicolumn{2}{c}{\textbf{PCC} (↑)} \\
    \cmidrule(lr){4-5}
      &  &  & \multicolumn{1}{c}{\textbf{Spot}} & \multicolumn{1}{c}{\textbf{Gene}} \\
    \midrule
    UNI    & 0.412  & 0.398   & 0.554 (0.249)          &  0.663 (0.111)\\
    STFlow & 0.387                    & 0.381                    & 0.581 (0.103)                  & 0.700 (0.104) \\
    ASIGN 3D & 0.689                    & 0.71                    & 0.362 (0.178)                  & 0.442 (0.013)\\
    
    \rowcolor{cyan!10}  Ours &
\textbf{\num{0.330}} &
\textbf{\num{0.343}} &
\textbf{\num{0.598(0.105)}} &
\textbf{\num{0.726(0.105)}} \\
    \bottomrule
  \end{tabular}
  }
\end{table}

% (optional, but nice for aligning units)
% \usepackage{siunitx}  % in your preamble

\begin{table}[t]
    \centering
    \scriptsize
    \setlength{\tabcolsep}{6pt}
    \renewcommand{\arraystretch}{1.15}
    \caption{Computational complexity on in-house embryo dataset: number of spots ($N$), forward-pass FLOPs, and peak GPU memory.}
    \vspace{-2mm}
    \begin{tabular}{
      l
      S[table-format=6.0]
      S[table-format=3.2]
      S[table-format=2.1]
    }
        \toprule
        \textbf{Model} & \multicolumn{1}{c}{\textbf{\# Spots ($N$)}} & \textbf{GFLOPs} & \multicolumn{1}{c}{\textbf{Memory [GB]}} \\
        \midrule
        ASIGN (3D)  & 327532 & {6032.08} & 79.8 \\
        Our method (Train) & 327532 & 333.21 & 26.9 \\
        Our method (Inference) & 327532 & 114.9 & 3.7 \\
        \bottomrule
    \end{tabular}
    \label{tab:complexity_spots_flops_mem}
\end{table}

\begin{table}[t]
  \centering
  \caption{
    Gene-level performance (median per-gene MSE, mean per-gene PCC) on the even-slice hold-out split for different ablation settings on the in-house embryo dataset.
    For each panel size, the best value is shown in \textbf{bold}.
    Lower is better for MSE/MAE; higher is better for PCC. Rows are cumulative: Vanilla STFlow, then the learned prior alone, then ControlNet, then the cross-slide branch.
    {--}: MAE was not recorded for this run.
  }
  \label{tab:ablation_component}
  \scriptsize
  \begin{tabular}{
    @{}
    l
    S[table-format=3.0]
    S[table-format=0.3]
    S[table-format=0.3]
    S[table-format=0.3]
    @{}
  }
    \toprule
    \textbf{Regime} & {\textbf{Genes}} & {\textbf{MSE} $\downarrow$} & {\textbf{MAE} $\downarrow$} & {\textbf{PCC(GENE)} $\uparrow$} \\
    \midrule
    \multirow{4}{*}{Vanilla STFlow}
      & 50  & 0.160 & 0.107 & 0.693 \\
      & 100 & 0.177 & 0.120 & 0.688 \\
      & 200 & 0.176 & 0.119 & 0.673 \\
      & 500 & 0.164 & \textbf{0.124} & 0.646 \\
    \midrule
    \multirow{4}{2.5cm}{Learned Prior}
      & 50  & 0.181 & {--} & 0.691 \\
      & 100 & 0.175 & {--} & 0.690 \\
      & 200 & 0.161 & {--} & 0.680 \\
      & 500 & 0.164 & 0.134 & 0.642 \\
    \midrule
    \multirow{4}{2.5cm}{ControlNet \\ +~Learned Prior}
      & 50  & \textbf{0.154} & 0.096 & \textbf{0.701} \\
      & 100 & 0.169 & 0.117 & 0.692 \\
      & 200 & 0.155 & 0.108 & 0.694 \\
      & 500 & 0.164 & 0.127 & 0.651 \\
    \midrule
    \multirow{4}{2.5cm}{ControlNet \\ +~Learned Prior \\ +~Cross Slide Neighbors}
      & 50  & 0.155 & \textbf{0.092} & 0.695 \\
      & 100 & \textbf{0.158} & \textbf{0.112} & \textbf{0.697} \\
      & 200 & \textbf{0.147} & \textbf{0.108} & \textbf{0.698} \\
      & 500 & \textbf{0.162} & 0.128 & \textbf{0.654} \\
    \bottomrule
  \end{tabular}
\end{table}

\paragraph{Effect of the learned prior.}
The learned ZINB prior has a mixed but generally positive effect relative to Vanilla STFlow: for the 100- and 200-gene panels it lowers MSE while slightly improving PCC (e.g., at 200 genes MSE drops from 0.176 to 0.161 and PCC rises from 0.673 to 0.680), whereas at 50 and 500 genes it does not consistently outperform the baseline, suggesting prior learning helps most for medium-sized panels.

\paragraph{Effect of ControlNet conditioning.}
Adding ControlNet on top of the learned prior improves PCC at every panel size and MSE at 50--200 genes over both Vanilla STFlow and the prior-only model. At 50 genes, PCC increases from 0.693 (vanilla) and 0.691 (prior only) to 0.701 while MSE decreases to 0.154, the best among all regimes at this panel size; similar trends hold at 100 and 200 genes (PCC $\approx0.69$--$0.70$ with reduced MSE/MAE), and at 500 genes it improves PCC (0.651 vs.\ 0.646) with unchanged MSE. Qualitatively, for gene AHSP our ControlNet- and cross-slide-conditioned model recovers high-expression clusters and fine-grained structure more faithfully than STFlow across flow steps (\cref{fig:controlnet}), visually reinforcing these quantitative gains (Table~\ref{tab:ablation_component}).

\paragraph{Effect of the cross-slide branch.}
Our full model, augmenting ControlNet with the cross-slide branch, delivers the strongest overall performance. At 100 and 200 genes it is best or tied on all three metrics (e.g., at 200 genes MSE improves from 0.155 to 0.147 and PCC from 0.694 to 0.698). At 500 genes, where all methods degrade slightly, it still improves over the prior-only variant (MAE 0.128 vs.\ 0.134; PCC 0.654 vs.\ 0.642) and attains the highest PCC among all regimes, indicating that cross-slide cues add complementary long-range context beyond what ControlNet and the learned prior capture alone.

\subsection{Additional baseline (TRIPLEX)}
\label{app:rebuttal-extra}

\paragraph{Comparison with TRIPLEX.}
We additionally re-train the per-section predictor TRIPLEX~\cite{Chung2024TRIPLEX} under our protocol on the two public datasets. \method substantially outperforms TRIPLEX on both PCC-gene and MSE (\cref{tab:triplex}); TRIPLEX's per-section predictive formulation cannot exploit the cross-section $z{\pm}1$ context that is central to our method.

\begin{table}[t]
\centering
\small
\setlength{\tabcolsep}{8pt}
\caption{Comparison with the per-section predictor TRIPLEX~\cite{Chung2024TRIPLEX}, re-trained under our protocol on the two public datasets (best in bold).}
\label{tab:triplex}
\begin{tabular}{llcc}
\toprule
Dataset & Method & PCC-gene\,$\uparrow$ & MSE\,$\downarrow$ \\
\midrule
\multirow{2}{*}{HER2}    & TRIPLEX        & $0.335{\pm}0.07$          & $0.486$ \\
                         & \method (ours) & $\mathbf{0.638}{\pm}0.15$ & $\mathbf{0.455}$ \\
\midrule
\multirow{2}{*}{ST-Data} & TRIPLEX        & $0.311{\pm}0.07$          & $0.980$ \\
                         & \method (ours) & $\mathbf{0.612}{\pm}0.13$ & $\mathbf{0.364}$ \\
\bottomrule
\end{tabular}
\end{table}

\subsection{Interpolation baseline}
\label{app:interpolation}
A simple alternative to imputation is to copy measured expression from the nearest profiled sections. For each spot of a held-out section we take the spot nearest to $(a_i,b_i)$ on the closest profiled section below and above and average the two (\emph{nearest spot}), or average the 4 nearest spots on each of the two sections (\emph{4 nearest spots}). \Cref{tab:interp} compares this with STFlow and \method on the in-house embryo even-slice hold-out, where every held-out section has profiled neighbours at $\Delta z{=}1$ and interpolation is a strong baseline. \method improves PCC-gene over the 4-nearest-spot interpolation by $+0.009$ ($68\%$ of genes better, $p{=}2{\times}10^{-5}$, paired Wilcoxon test across genes).

\begin{table}[t]
\centering
\small
\setlength{\tabcolsep}{6pt}
\caption{Interpolation baseline versus STFlow and \method\ on the in-house embryo (even-slice hold-out, 150-gene panel, mean${\pm}$sd over 3 seeds, last checkpoint). STFlow and \method\ are re-runs and differ slightly from the single runs in \cref{tab:three_datasets_metrics}. Interpolation copies measured expression from the closest profiled sections below and above (nearest spot by $(a,b)$, or mean of the 4 nearest spots, averaged over the two sections). Best in bold.}
\label{tab:interp}
\begin{tabular}{lcc}
\toprule
Method & PCC-gene\,$\uparrow$ & PCC-spot\,$\uparrow$ \\
\midrule
Interpolation (nearest spot)    & $0.647$ & $0.521$ \\
Interpolation (4 nearest spots) & $0.688$ & $0.582$ \\
STFlow                          & $0.650{\pm}0.004$ & $0.556{\pm}0.003$ \\
\method (ours)                  & $\mathbf{0.697}{\pm}0.001$ & $\mathbf{0.583}{\pm}0.003$ \\
\bottomrule
\end{tabular}
\end{table}

\end{document}